\documentclass[runningheads]{llncs}

\usepackage[final,year=2026,ID=870]{accv}

\usepackage{accvabbrv}

\usepackage{graphicx}
\usepackage{booktabs}

\usepackage{siunitx}                 
\usepackage{multirow}                
\usepackage{placeins}                
\usepackage{enumitem}
\usepackage{tikz}
\usepackage{pgfplots}
\pgfplotsset{compat=1.18}
\usepgfplotslibrary{units}
\usetikzlibrary{pgfplots.groupplots}
\usepackage[mode=image]{standalone}   
\usepackage[percent]{overpic}            
\usepackage{wrapfig}    

\definecolor{avatColor}{HTML}{5C8A8A}
\definecolor{fastColor}{HTML}{C98A7D}
\definecolor{oursColor}{HTML}{0000FF}

\providecommand{\Description}[1]{}

\usepackage[pagebackref,breaklinks,colorlinks,citecolor=accvblue]{hyperref}
\usepackage{orcidlink}

\begin{document}

\title{GHARP: Real-time Gaussian Head Animation from Large-scale Reconstruction Prior}

\titlerunning{GHARP: Real-time Gaussian Head Animation}

\author{Ali Benlalah\textsuperscript{$\dagger$} \and
      Sepehr Johari\textsuperscript{$\dagger$} \and
      Patricia Vitoria\textsuperscript{$\dagger$} \and
      Armin Kappeler\textsuperscript{$\dagger$} \and \linebreak
      Artem Sevastopolsky \and
      Alexander Jung \and
      Gabriele Fanelli \and
      Kevin Mader \and \linebreak
      Manuel Breitenstein \and
      Claudia Pl\"uss \and
      Jan R\"uegg \and
      Simon Biland \and \linebreak
      Thomas Etterlin \and
      Dmitry Kostiaev \and
      Mathias Deschler \and
      Brian Amberg \and
      Sebastian Martin}
\authorrunning{A. Benlalah et al.}
\institute{Apple \\
    \email{\{abenlalah,johari,p\_vitoriacarrera,akappeler\}@apple.com}
}

\maketitle
\begingroup
\renewcommand{\thefootnote}{}   
\footnotetext{$^{\dagger}$ These authors contributed equally to this research.}
\endgroup

\vspace{-10pt}

\begin{abstract}
We present GHARP (Real-time \textbf{G}aussian \textbf{H}ead \textbf{A}nimation from Large-scale
\textbf{R}econstruction \textbf{P}rior), a method that animates 3D human heads in real time from a few input images of a subject and a driving expression signal. We decouple the problem into an identity stage that builds a representation of the subject's geometry and appearance offline, and an animation stage that predicts expression-dependent residuals on top of it at runtime. This separation offers a favorable trade-off with respect to fidelity, quality and runtime: the identity stage can be expensive while the animation stage runs a lightweight network, optimized for mobile devices. Our method performs animation in a semantically structured latent space of a pretrained reconstruction model, where expression changes remain spatially contained, making residual prediction efficient. This reconstruction prior provides a consistent spatial layout, allowing fusion of multiple input views into a compact, fixed-size canonical Gaussian representation. While this two-stage design improves the runtime-quality trade-off, it still inherits a problem common to all expression-driven avatar methods: expression codes describe only the face and thus omit body pose and clothing position, making these regions underspecified in the input. The animation network faces an ill-posed mapping and resorts to averaging over conflicting body appearances, producing blur and temporal flicker. We address this with a body alignment network that learns to align the person's body in the target image with the input reference images, removing the ambiguity from the training signal. Our method achieves state-of-the-art quality on the Ava-256 benchmark while running up to $13\times$ faster on an A100 GPU with $8\times$ fewer Gaussians.

\keywords{Fast Avatar Generation \and Efficient On-Device Rendering \and Parametric Head Model \and Real-Time Avatar Animation}
\end{abstract}
\vspace{-20pt}

\begin{figure}[t]
    \raisebox{7pt}{\includegraphics[width=0.6\columnwidth, trim=0 0 0 0, clip]{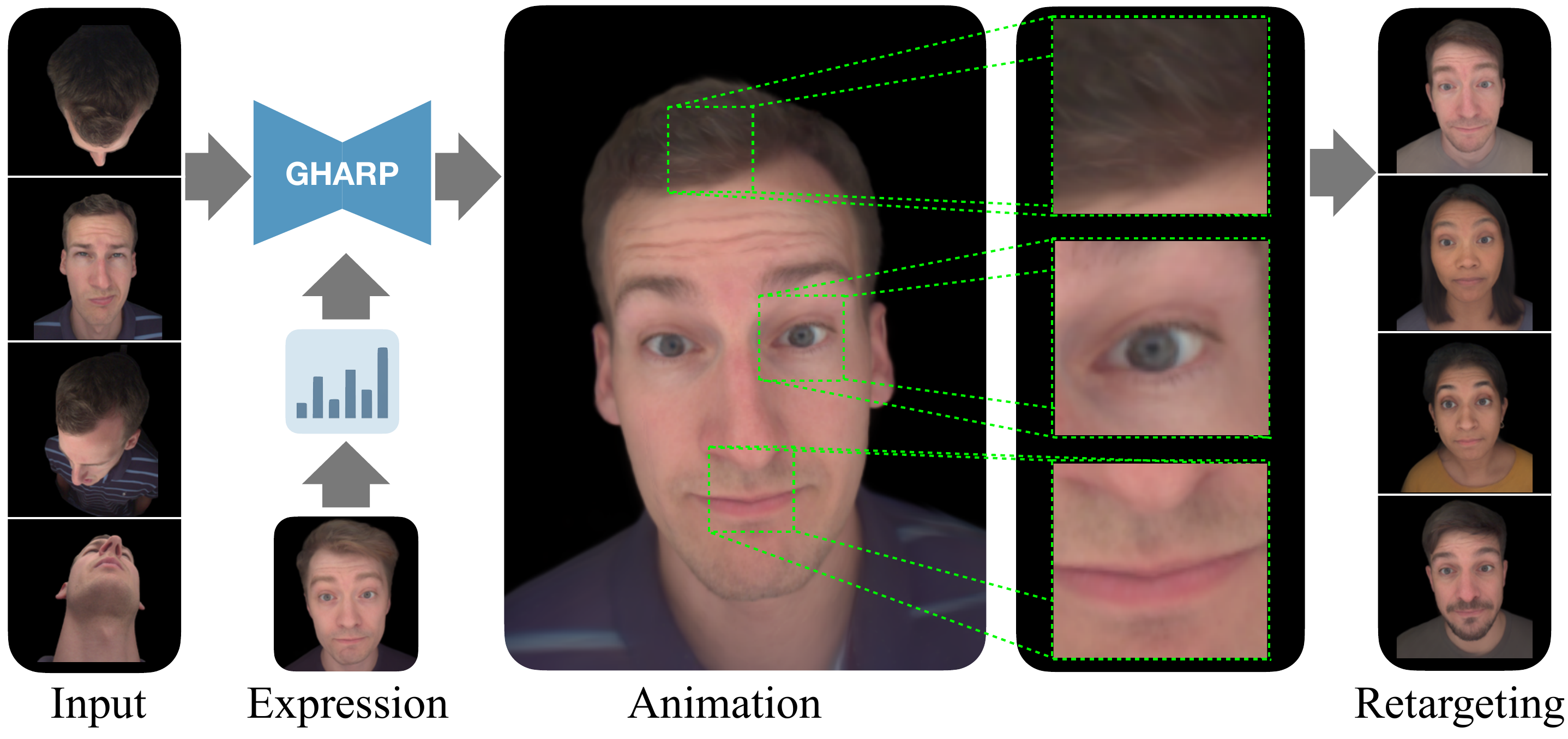}}
    \hfill
    \includegraphics[width=0.4\columnwidth]{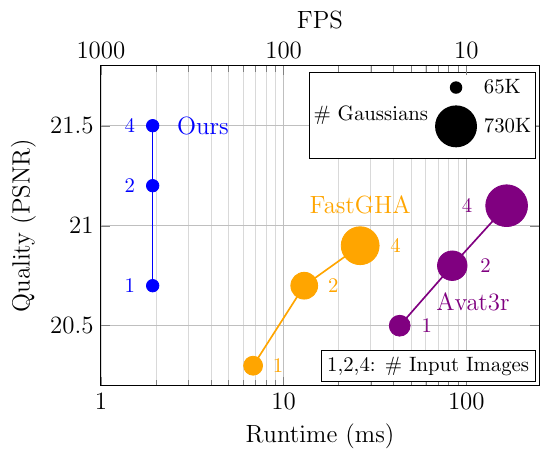}
      \vspace{-20pt}
    \caption{Left: GHARP animates a subject from a set of input images and an expression code. Additionally, GHARP enables retargeting of the same expression to new subjects with state-of-the-art quality. Right: runtime on A100 vs. reconstruction quality on Ava-256 for 1, 2 and 4 input images. GHARP is an order of magnitude faster. }
    \vspace{-20pt}
    \label{fig:teaser}
\end{figure}

\section{Introduction}

Creating photorealistic, real-time animated head avatars is a central challenge in computer vision and graphics, with applications in telepresence, VR, and digital communication. Recent advances in 3D Gaussian Splatting~\cite{kerbl3Dgaussians} have made real-time, high-quality rendering of 3D scenes practical, spurring a growing body of work on Gaussian-based head avatars.


Per-subject optimization methods~\cite{qian2024gaussianavatars,li2024uravatar,aneja2025scaffoldavatarhighfidelitygaussianavatars} achieve impressive quality by fitting a Gaussian Splatting model to dense multi-view captures, but require expensive light-stage hardware and hours of per-subject optimization, making them impractical for scalable deployment. Feed-forward methods remove this requirement: a neural network takes a few input images of a subject paired with an expression code and directly predicts an animated avatar. Avat3r~\cite{kirschstein2025avat3r} and FastGHA~\cite{ji2026fastgha} demonstrated high-quality feed-forward reconstruction and animation, yet both predict pixel-aligned Gaussians for each input view, causing the total Gaussian count and inference cost to grow linearly with the number of input images, precluding deployment on resource-constrained devices.


These limitations share a common root: jointly learning reconstruction
and animation couples the cost of animation to the reconstruction complexity. In this work, we show that performing animation in a structured latent space of a pretrained reconstruction model decouples these two processes while also providing a strong prior for animation. We use HeadsUp~\cite{headsup} as a reconstruction prior to encode input images into 2D latents. A convolutional decoder then maps these latents into Gaussian UV maps, i.e. UV-parameterized Gaussians anchored to a neutral head template. The convolutional nature of the decoder ensures that a 2D position in the latent maps to a fixed facial region. Therefore, latents from input images with different expressions are semantically aligned. This structure allows us to model a change in expression as a localized delta rather than a global transformation. However, running the decoder per frame would be too costly. 
Instead, we build a canonical Gaussian UV map once, then predict a residual on top at animation time. 
Since the number of Gaussians remains constant, this decouples the animation cost from both the reconstruction model's complexity and the number of input views.

This residual formulation benefits from a canonical representation that is as close to the target as possible, thereby reducing the variance of the residuals predicted at runtime. 
We construct this canonical representation by fusing multiple input views in the latent space into a compact representation that captures the subject's face geometry and appearance across viewpoints and expressions. This representation is then decoded into the canonical Gaussian UV map.
To further model person-specific facial motion across expressions, we introduce a deformation latent that encodes subject-specific facial dynamics. Both the canonical representation and the deformation latent are pre-computed once per subject. Consequently, runtime inference requires only a lightweight convolutional network that takes as input the canonical Gaussian UV map, the deformation latents and a target expression code to predict the residual.


  
Body pose and clothing may also vary across input images independently of facial expression. Since the expression code does not explain these variations, the network averages over conflicting body appearances, which manifests as temporal jittering during animation. To resolve this ambiguity, we introduce a \emph{body alignment network} (BAN) that composites supervision targets in the shared UV-aligned latent space, taking the facial expression from the target and the body appearance from a reference image.
To summarize, our main contributions are:
\begin{enumerate}[topsep=0pt]
\item A \textbf{real-time, animation-focused head avatar method,} based on a decoupled identity and animation design. With the identity network running only once per subject, the approach achieves a faster runtime by using a lightweight animation network that animates the avatar at a constant cost, independent of the number of input views.

\item A \textbf{body alignment network (BAN)} that resolves the ill-posed animation of body regions not described by the expression signal, improving temporal stability at animation time.

\item We show that the UV-parameterized latent space of a pretrained face reconstruction model is a \textbf{structured prior for avatar animation}: its spatial alignment across different input images enables identity construction, animation, and body alignment to all operate within a single shared representation.
\end{enumerate}
Our experiments show state-of-the-art results at a fraction of the cost: on an A100, we are up to 13$\times$ faster than FastGHA~\cite{ji2026fastgha} and 86$\times$ faster than Avat3r~\cite{kirschstein2025avat3r}; on an iPhone 17 Pro, we are 31$\times$ faster than FastGHA with 8$\times$ fewer Gaussians, achieving 522 FPS on an A100 and 690 FPS on an iPhone 17 Pro.
\vspace{-10pt}

\section{Related Work}



\begin{figure*}[t!]
  \includegraphics[width=1.0\textwidth, trim=0 0 0 0, clip]{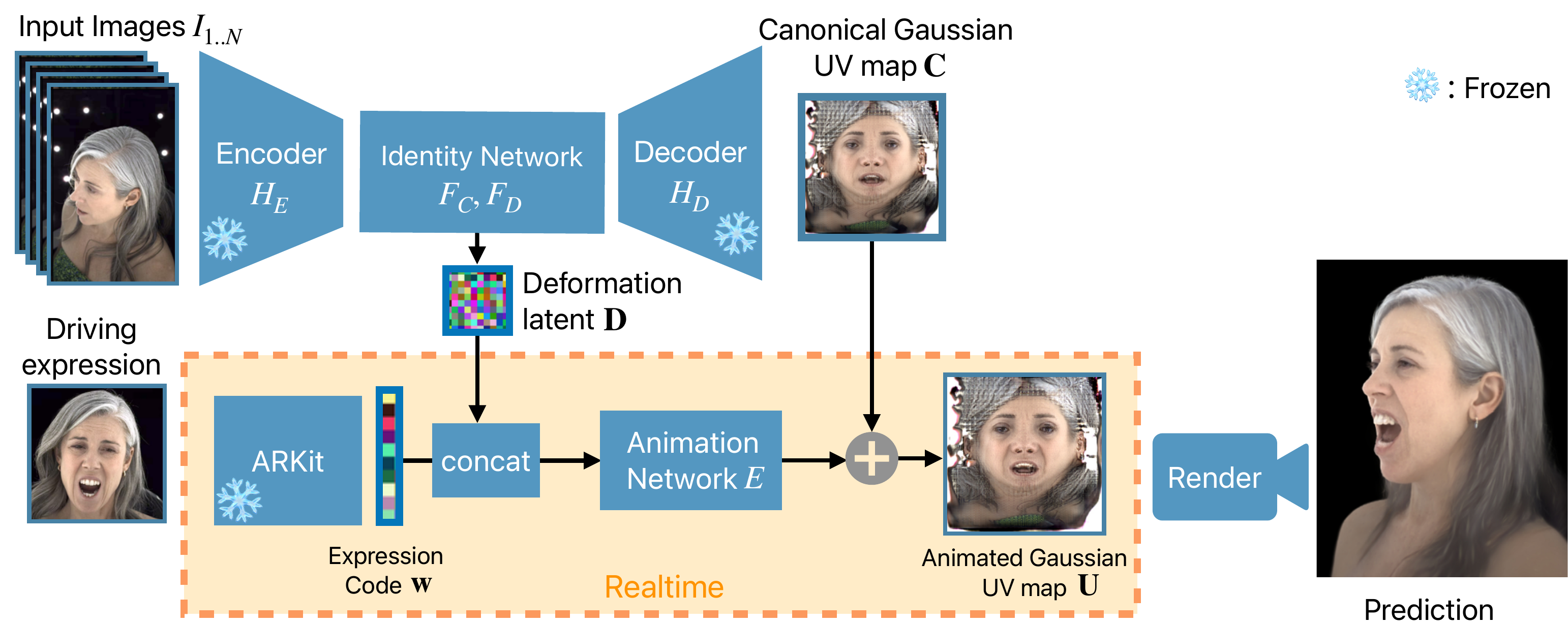}
    \vspace{-20pt}
  \caption{\textbf{Architecture.} Input images are encoded into per-view latents that are aggregated by $F_C$ into a fixed-size canonical Gaussian UV map $\mathbf{C}$ and by $F_D$ into a deformation latent $\mathbf{D}$, both independent of the number of input views $N$. At runtime, only the lightweight animation network operates on $\mathbf{C}$, $\mathbf{D}$, and an expression code $\mathbf{w}$ to produce the animated Gaussian UV map.}
 \vspace{-20pt}
  \label{fig:architecture}
\end{figure*}

\noindent \textbf{Subject-Specific Avatar Creation.} 
Pioneering works on real-time head avatars used differentiable 3D Morphable Face Models (3DMMs)~\cite{3dmm,paysan20093d,li2017flame,egger20203dmorphablefacemodels,guo2021fastaccuratestable3d} to enable efficient optimization frameworks for both tracking and rendering of faces; the relatively low-dimensionality of these models however prevents faithful representation of complex details in geometry and appearance such as hair or clothing. 
Volumetric methods like NeRF-based approaches~\cite{park2021nerfies,athar2022rignerf,zielonka2023instant,Xu_2023,Gao2022nerfblendshape} have gained popularity recently as they allow for more complex modeling and photorealistic results at the cost of expensive, per-scene optimization and, in most cases, rendering speeds far below real-time.
Alternative representations including deformable point clouds~\cite{zheng2023pointavatar}, implicit surfaces~\cite{zheng2022imavatar,yenamandra2020i3dmmdeepimplicit3d}, and mesh-based neural rendering~\cite{grassal2022neural} improved efficiency or geometric coverage but still fall short of real-time performance. Neural Parametric Head Models~\cite{giebenhain2024mononphm,zanfir2023phomohimplicitphotorealistic3d,Giebenhain_2024} learn non-linear expressive shape and appearance latent spaces, yet remain too slow for real-time use and require per-subject optimization.
The highest-fidelity head avatars to date~\cite{lombardi2018deep,ma2021pixelcodec,saito2024rgca,li2024uravatar,li2025tega} are obtained from dense, multi-view light-stage captures of a person.
A fast-growing body of work~\cite{Ma_2024,qian2024gaussianavatars,shao2024splattingavatar,aneja2025scaffoldavatarhighfidelitygaussianavatars,teotia2024gaussianheads,dhamo2024headgas,tang2025gaf,zielonka2025gaussianeigenmodelshuman,sun2025fine,xu2025gphm,xu2024gaussian,wang20253d,mendiratta2025grmm,zhang2025hravatar} has in recent years embraced 3D Gaussian Splatting as the representation of choice for head avatars, achieving real-time rendering with high visual fidelity, but at the cost of per-subject optimization that prevents generalization to unseen identities. 
In contrast, our method employs a feed-forward architecture, largely inspired by~\cite{headsup}, that generalizes to unseen subjects without per-subject optimization.

\noindent \textbf{Feed-forward Avatar Creation.}    

\textit{Animation in 2D.}
A parallel line of work performs head reenactment directly in 2D or with shallow 3D proxies~\cite{drobyshev2023megaportraitsoneshotmegapixelneural,khakhulin2022realisticoneshotmeshbasedhead,guo2025liveportraitefficientportraitanimation,tran2024voodooxpexpressiveoneshot,ma2023cvtheadoneshotcontrollablehead,ye2024real3dportraitoneshotrealistic3d,deng2024portrait4d,ren2025egavatar,deng2024portrait4dv2,ma2023otavatar}, achieving fast and generalizable animation, but without true viewpoint control or explicit 3D geometry. In contrast, our method produces explicit 3D geometry with true viewpoint control --- a combination lacking in 2D reenactment approaches.

\textit{Animation in 3D.}
Recent feed-forward methods have shown that cross-subject generalization and sparse-input reconstruction are achievable from multiple images~\cite{kirschstein2025avat3r,ji2026fastgha,chu2024gpavatar,wu2026uika,zheng2025headgap} or even a single image~\cite{he2025lam,chu2024gagavatar,lyu2024facelift,kant2025pippo}. However, most approaches either struggle with real-time animation or rely on per-pixel Gaussian representations~\cite{kirschstein2025avat3r,ji2026fastgha} that compromise view and temporal consistency, with rendering cost and memory that grow linearly with the number of input views. In contrast, we introduce a fixed-size identity representation that decouples animation cost from the input count and a body alignment network that maintains consistent geometry across views and time.
  
\noindent\textbf{Accelerated Avatars.}
Several methods specifically target real-time or on-device inference, mainly focused on subject-specific settings~\cite{xiang2024flashavatar,Duan_2023,wang2024lightavatarefficientheadavatar,liang2025fastavatar,raina2025prismavatarrealtimeanimated3d,yan2025architectheadcontinuouslevelcontrol,medin2026legacyavatarsvolumetricfaceavatars,jiang2025mobileportrait}. 
While these works demonstrate that speed and quality can coexist, they require per-identity optimization and thus cannot scale to multiple unseen subjects without retraining. 
Some of the very recent works improve along both quality and speed axes while operating in the generalized setting, such as~\cite{he2025lam,wu2026uika,kirschstein2026flexavatar,prinzler2026feed,li2026large}, with performance ranging from $\sim$2 FPS for~\cite{prinzler2026feed} to a more attractive 30 FPS~\cite{kirschstein2026flexavatar} on a consumer GPU to very competitive FPS numbers of~\cite{he2025lam,wu2026uika,li2026large}.
The latter group of methods, however, relies on linear blend skinning (LBS) for animation --- a reasonable choice for full-body modeling but limiting facial expression rigging to linear motion of an underlying coarse parametric model.
Our method offers a lightweight feed-forward network that generalizes across subjects and produces sophisticated, non-linear facial expression animations while maintaining real-time animation cost.
\vspace{-10pt}

\section{Method}

\vspace{-5pt}

An overview of our method can be seen in Fig. \ref{fig:architecture}. We use a learned prior based on an encoder $H_E$ and decoder $H_D$ architecture. The encoder $H_E$ compresses calibrated input views into a compact UV-aligned 2D latent representation $\mathbf{Z} \in \mathbb{R}^{32 \times 32 \times 512}$. The decoder $H_D$ maps latents to UV-parameterized 3D Gaussians anchored to a neutral head mesh template.
We refer to it as a Gaussian UV map with spatial dimensions $H_{UV}$ and $W_{UV}$. 
Each Gaussian is characterized by
a position $\boldsymbol{\mu} \in \mathbb{R}^{3}$,
a scale $\mathbf{s} \in \mathbb{R}^{3}_{+}$,
a unit quaternion rotation $\mathbf{q} \in \mathbb{R}^{4}$,
an opacity $\mathbb{\alpha} \in [0,1]$,
and spherical harmonic color coefficients $\mathbf{k} \in \mathbb{R}^{12}$ (SH degree 1),
for a total of $d_G = 3{+}3{+}4{+}1{+}12$ attributes per Gaussian.
Our method takes $N$ input images of a subject $\{\mathbf{I}_1, \ldots, \mathbf{I}_N\, \mid \mathbf{I}_i \in \mathbb{R}^{H \times W \times 3} \}$ and an expression code $\mathbf{w}$ and produces an animated avatar represented as a Gaussian UV map $\mathbf{U} \in \mathbb{R}^{H_{UV} \times W_{UV} \times d_G}$.

We formulate animation as residual learning in the canonical Gaussian space. A change of expression between two frames corresponds to a localized delta on top of the Gaussian attributes. This residual deforms non-rigid regions of the face, such as the cheeks, mouth, and eyelids, while keeping attributes like skin texture and pores almost intact.
Let $\mathbf{C} \in \mathbb{R}^{H_{UV} \times W_{UV} \times d_G}$ denote a canonical Gaussian UV map, a set of splats that encode the subject's static appearance and geometry. We learn a lightweight animation network $E$ such that:
\begin{equation}
  \mathbf{U} = E(\mathbf{w}, \mathbf{D}) + \mathbf{C},
  \label{eq:residual}
\end{equation}
where $\mathbf{w}$ is an expression code and $\mathbf{D}$ denotes the deformation latent.
 
 While $\mathbf{C}$ is a learned canonical representation that combines identity information from the input images (Sec.\ref{subsec:canonical-identity}), $\mathbf{D}$ encodes face deformations across expressions. This decomposition minimizes the workload of $E$ by offloading as much information as possible into $\mathbf{C}$. We detail the architecture of $E$ in Sec.\ref{subsec:animation-network}.

\subsection{Offline Identity Subspace Learning}
\label{subsec:canonical-identity}

The canonical Gaussian UV map $\mathbf{C}$ and the deformation latent $\mathbf{D}$ together form a subject's identity subspace, a compact representation that captures both static appearance and expression-dependent motion. 
During training, the canonical map is shared across all of a subject's expressions and therefore captures expression-invariant identity, while the deformation latent, trained end-to-end with the animation network, captures expression-dependent motion.
At inference time, both are constructed from the input images once per subject offline. The canonical path fuses appearance information from the input images into $\mathbf{C}$, and the deformation path encodes how the subject's face deforms across expressions. 
We use a reconstruction objective that anchors the canonical map during training~(Sec.~\ref{subsec:passthrough}).

\begin{figure}[t]
  \centering
  \begin{tikzpicture}
    \node[anchor=south west, inner sep=0] (img)
      {\includegraphics[width=\columnwidth, trim=0 0 0 0, clip]{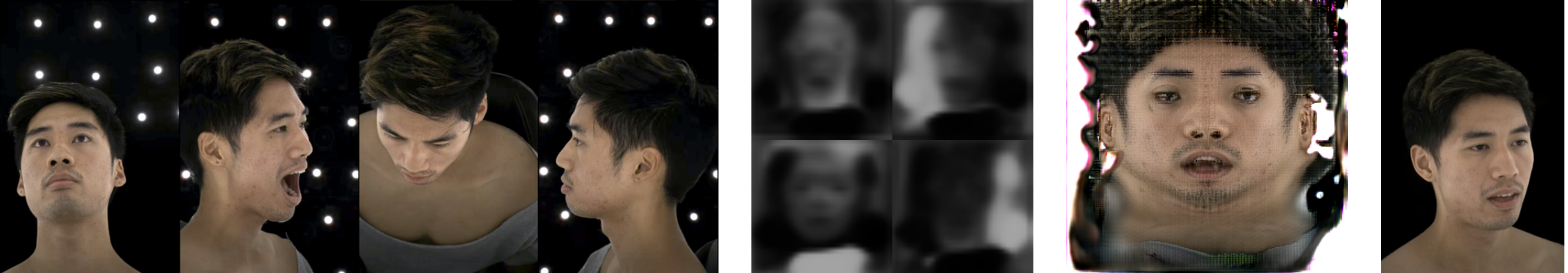}};
    \begin{scope}[every node/.style={anchor=base west, font=\scriptsize, yshift=2pt}]
      \node at ([xshift=0.17\columnwidth]img.north west) {Input Images};
      \node at ([xshift=0.47\columnwidth]img.north west) {Blending Weights};
      \node at ([xshift=0.69\columnwidth]img.north west) {Canonical UV};
      \node at ([xshift=0.85\columnwidth]img.north west) {Canonical Render};
    \end{scope}
  \end{tikzpicture}
  \caption{\textbf{Canonical latent blending.} The canonical Gaussian UV map consistently combines input view details (e.g. teeth, hair) with high fidelity.}
  \vspace{-20pt}
  \label{fig:canonical_blending}
\end{figure}

\noindent \textbf{Canonical Gaussian UV Map.}
The quality of the residual decomposition in Eq.~\eqref{eq:residual} depends on how well $\mathbf{C}$ captures the subject's identity:
a more complete canonical representation yields smaller residuals, which simplifies the task of $E$ and results in higher-fidelity animation. We encode each input image into a latent $\mathbf{Z_i} = H_E(\mathbf{I}_i)$ yielding a set of latents $\{\mathbf{Z_i}\}_{i=1}^{N}$. These latents are concatenated with their corresponding 6D Plücker~\cite{sitzmann2021light} embeddings to encode camera geometry and are fed to a canonical fusion network $F_{C}$ that predicts per-view spatial blending weights $\{b_i\}_{i=1}^{N}$ and a correction term $\boldsymbol{\delta}_c$. The resulting latents $\mathbf{C}_{\ell}$ are then decoded with $H_D$ to obtain the canonical Gaussian UV map $\mathbf{C}$:
\begin{equation}
    \mathbf{C} = H_D(\mathbf{C}_{\ell}),
    \quad 
    \mathbf{C}_{\ell} = \sum_{i=1}^{N} \sigma(b_i) \cdot \mathbf{Z}_{i} + \boldsymbol{\delta}_c
  \label{eq:canonical}
\end{equation}
where $\sigma$ is the sigmoid function. By operating in the latent space, the blending learns to optimally combine geometry, appearance, and fine details from all input views (see Fig. \ref{fig:canonical_blending}), while $\boldsymbol{\delta}_c$ allows small adjustments beyond the span of the input latents, such that the resulting $\mathbf{C}$ maximally offloads work from the animation network.

\noindent  \textbf{Deformation Latent.} While $\mathbf{C}$ encodes the subject's static identity by fusing all input views, it does not encode expression-dependent motion, i.e.,\ how this subject's face deforms across expressions. To supply this missing information, we first decode each per-view latent $\mathbf{Z_i}$ to $\mathbf{G}_i = H_D(\mathbf{Z_i}) $. Then, a convolutional deformation encoder $F_D$ processes these maps together with $\mathbf{C}$ from Eq.~\eqref{eq:canonical} to produce the deformation latent:
\begin{equation}
  \mathbf{D} = F_{D}\!\left(\{\mathbf{G}_i\}_{i=1}^{N},\; \mathrm{sg}(\mathbf{C})\right),
  \label{eq:deformation}
\end{equation}
where $\mathrm{sg}(\cdot)$ denotes the stop-gradient operator. $F_D$ is a U-Net that aggregates per-view Gaussian UV maps into a fixed-size tensor $\mathbf{D} \in \mathbb{R}^{64 \times 64 \times 96}$. We detach $\mathbf{C}$ from the computation graph of $F_D$ so that the canonical path receives no gradients through the deformation encoder. This drives the canonical map to specialize in expression-agnostic appearance and geometry, while the deformation latent, trained end-to-end with the animation network, learns to encode expression-dependent deformation. We refer to Supp. Mat.~\ref{app:architecture_details} for architecture details.


\subsection{Animation Network}
\label{subsec:animation-network}

The animation network $E$ in Eq.~(\ref{eq:residual}) takes as input $\mathbf{w}$ and $\mathbf{D}$, and outputs a residual for each Gaussian parameter in the canonical UV map. The expression code $\mathbf{w} \in \mathbb{R}^{M}$, where $M$ is the dimension of the expression code, is processed by an MLP, then tiled and concatenated with $\mathbf{D}$ along the channel dimension. $E$ is implemented as a ResNet~\cite{he2016deep} decoder which processes this input to produce a residual delta map $E(\mathbf{w}, \mathbf{D}) \in \mathbb{R}^{H_{UV} \times W_{UV} \times d_G}$. This architecture is a natural choice for this decoder as it exploits the spatial structure of the UV layout while executing efficiently on GPU for real-time animation. We refer to Supp. Mat.~\ref{app:architecture_details} for architecture details.

\subsection{Passthrough Auxiliary Loss}
\label{subsec:passthrough}


To further encourage the identity network $F_C$ to preserve texture details and facial geometry, we introduce a passthrough loss that anchors $\mathbf{C}$ to the input reference images, providing a direct reconstruction signal. Given the latents $\mathbf{Z_i}$, a lightweight encoder-decoder network predicts small corrections $\Delta\mathbf{P}_i$ (details in Supp. Mat.~\ref{app:architecture_details}) on top of the canonical:
\begin{equation}
  \hat{\mathbf{G}}_{i} = \mathbf{C} + \mathbf{M} \odot \Delta\mathbf{P}_i,
  \label{eq:passthrough}
\end{equation}
where $\hat{\mathbf{G}}_{i}$ is the reconstructed Gaussian UV map for the corresponding input view, $\odot$ denotes element-wise multiplication.  $\mathbf{M}$ is a binary channel mask that restricts the correction to Gaussian positions and opacities only, zeroing out rotations, scales, and color coefficients. Together with the narrow bottleneck, this limits the correction capacity and encourages the canonical map to focus on the appearance attributes, rather than relying on per-view reconstruction.
 $\hat{\mathbf{I}}^{\text{pass}}_{i}$ is rendered from $\hat{\mathbf{G}}_{\mathrm{i}}$ and supervised against $I_i$ from the same viewpoint. 
The passthrough auxiliary loss is defined as:
\begin{equation}
\mathcal{L}_{\text{pass}} = \sum_{i=1}^{N} \lambda^\text{pass}_{1} \cdot \| \hat{\mathbf{I}}^{\text{pass}}_{i} - \mathbf{I}_{i} \|_1 +
\lambda^\text{pass}_{\text{lpips}} \cdot \text{LPIPS}(\hat{\mathbf{I}}^{\text{pass}}_{i},\, \mathbf{I}_{i}).
\end{equation}

\subsection{Training Losses}
\label{subsec:training-losses}

We supervise our network through differentiable rendering. We render the Gaussians in $\mathbf{U}$ from the same viewpoint as the target image $\mathbf{I}_{GT}$ to obtain image $\hat{\mathbf{I}}$. The total training loss for $F_C$, $F_D$ and $E$ is:
\begin{equation}
\mathcal{L} = \lambda_1 \mathcal{L}_1 + \lambda_{\text{ssim}} \mathcal{L}_{\text{ssim}} + \lambda_{\text{lpips}}
  \mathcal{L}_{\text{lpips}} + \lambda_{\text{region}} \mathcal{L}_{\text{region}} + \lambda_{\text{reg}} \mathcal{L}_{\text{reg}} + \mathcal{L}_{\text{pass}}.
  \label{eq:total_loss}
\end{equation}
$\mathcal{L}_1$, $\mathcal{L}_{\text{ssim}}$, and $\mathcal{L}_{\text{lpips}}$~\cite{zhang2018unreasonable} are standard image reconstruction losses applied between $\hat{\mathbf{I}}$ and $\mathbf{I}_{GT}$. $\mathcal{L}_{\text{reg}}$ is an opacity and scale regularization following~\cite{saito2024rgca}, and $\mathcal{L}_{\text{region}}$ applies reconstruction losses on the eye and mouth regions. More details can be found in the Supp. Mat.~\ref{app:losses_details}.
\vspace{-10pt}

\begin{figure*}[t]
  \centering
  \begin{tikzpicture}
    \node[anchor=south west, inner sep=0] (img)
      {\includegraphics[width=\linewidth]{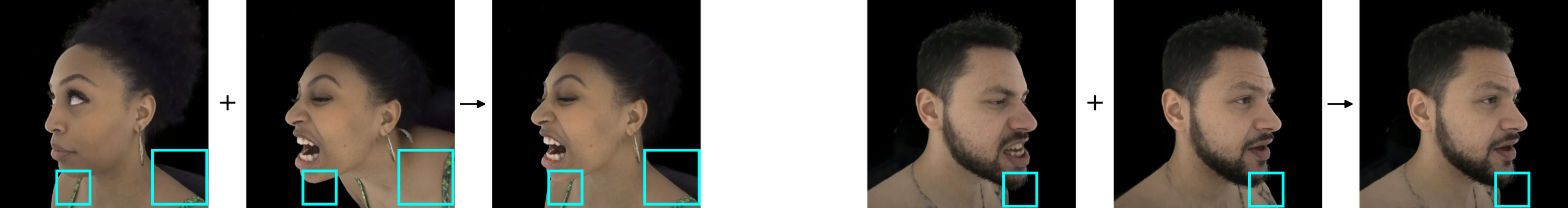}};
    \begin{scope}[
      every node/.style={anchor=base, font=\small, yshift=2pt}
    ]
      \node at ([xshift=0.07\linewidth]img.north west) {Reference};
      \node at ([xshift=0.22\linewidth]img.north west) {Target};
      \node at ([xshift=0.38\linewidth]img.north west) {Body-aligned};
      \node at ([xshift=0.62\linewidth]img.north west) {Reference};
      \node at ([xshift=0.77\linewidth]img.north west) {Target};
      \node at ([xshift=0.93\linewidth]img.north west) {Body-aligned};
    \end{scope}
  \end{tikzpicture}

  \vspace{-5pt}
  \noindent\textcolor{black!30}{\rule{\linewidth}{0.4pt}}

  \centering
    \includegraphics[width=\textwidth, trim=0 0 0 0,
  clip]{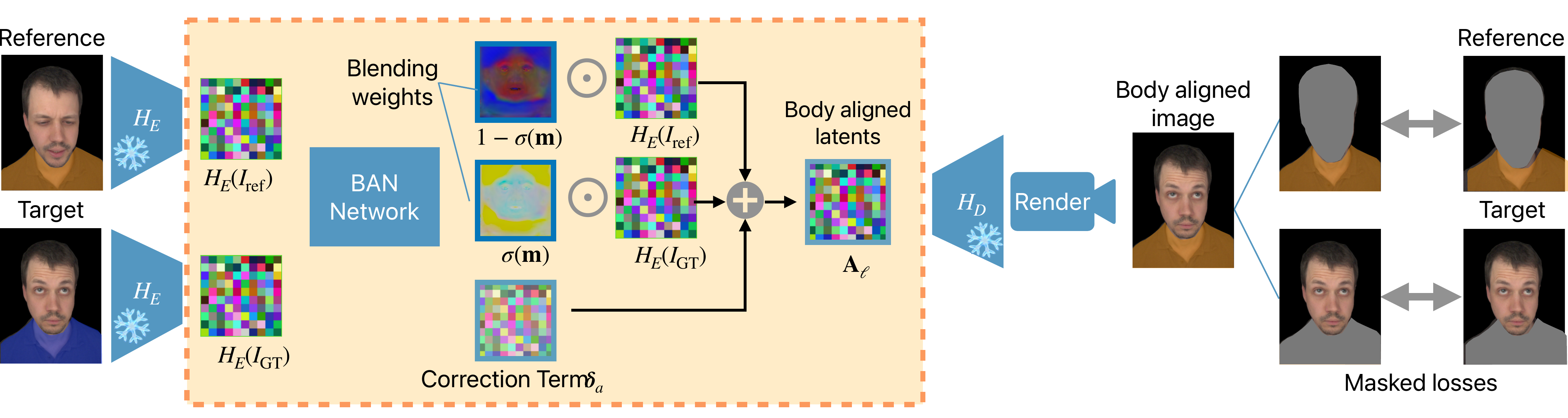}
    \caption{\textbf{Body Alignment Network (BAN).} \textit{Top row:} BAN takes the body pose from the reference image and the facial expression from the target image, producing a target that is consistent in body appearance with the reference (see cyan boxes).
    %
   \textit{Bottom row:} BAN Architecture. Reference and target images are encoded into a shared latent space and blended by predicted weights $\sigma(\mathbf{m})$ and a correction $\boldsymbol{\delta}_a$ into an aligned latent $\mathbf{A}_{\ell}$. Decoding yields the aligned Gaussian UV map $\mathbf{A}$, which carries the body from the reference and the face from the target.
   Face- and body-region image losses and an adversarial loss drive the blending (Supp.~Mat.~\ref{app:ban_details}).
   }
    \vspace{-14pt}
    \label{fig:gt_body_align_architecture}
\end{figure*}

\subsection{Body Alignment Network (BAN)}
\label{sec:body-alignment}
Body pose and clothing vary between input reference images independently of facial expression. Since the expression code cannot explain these variations, supervising against raw captures forces the network to average over conflicting body appearances, which manifests as temporal jitter at inference (Fig.~\ref{fig:body_alignment_vis}) and a loss of details. The body alignment network (BAN) addresses this by compositing face and body information in the latent space to produce aligned supervision targets.

As shown in Fig.~\ref{fig:gt_body_align_architecture}, BAN takes as input the reference latent $H_E(I_{\text{ref}})$, where $I_{\text{ref}}$ is one of the $N$ input images, and a target latent $H_E(I_{\text{GT}})$ and predicts spatial blending weights $m$ and a correction term $\delta_a$. The aligned latent is computed as:
\begin{equation}
  \mathbf{A}_{\ell} = \sigma(\mathbf{m}) \odot H_E(\mathbf{I}_{\mathrm{GT}})
    + (1 - \sigma(\mathbf{m})) \odot H_E(\mathbf{I}_{\mathrm{ref}})
    + \boldsymbol{\delta}_a,
  \label{eq:alignment}
\end{equation}
where $\boldsymbol{\delta}_a$ allows the output to deviate from a pure interpolation of the two inputs. The aligned latent is decoded with the frozen decoder to produce an aligned Gaussian UV map $\mathbf{A} = H_D(\mathbf{A}_{\ell})$. Compositing in the latent space keeps $\mathbf{A}$ on the decoder's data manifold, so any blended latent decodes to a plausible Gaussian head. Image-space compositing would instead introduce seams and lighting mismatches that propagate into the animation output. BAN is trained separately and frozen when producing supervision targets. 
During the animation network training, we replace the raw target $\mathbf{I}_{GT}$ in Eq.~(\ref{eq:total_loss}) with the rendering of $\mathbf{A}$ at the same viewpoint providing a consistent supervision signal. We refer to Supp. Mat.~\ref{app:ban_details} for training details.

\vspace{-5pt}



\section{Experiments}
\subsection{Datasets and Expression Codes}
\label{subsec:experiments-datasets}

\textbf{Internal10K Dataset.} We use an internal multi-view dataset of over \num{10}K subjects captured with a head-focused rig comprising \num{16} calibrated RGB cameras at a resolution of $1000\times750$ pixels. Each recording session includes a range of facial expressions and speech sequences. For training, we sample \num{100} expression-diverse frames per subject across all \num{10}K identities. Evaluation is performed on \num{100} frames from a held-out set of \num{50} subjects.

\noindent \textbf{Ava-256 Dataset.} 
We additionally finetune and evaluate on Ava-256~\cite{martinez2024codec}, a public multi-view dataset containing \num{256} subjects captured with \num{80} RGB cameras at $1024\times667$ resolution. We use the \num{4}TB variant of the dataset. Following the protocol of Avat3r~\cite{kirschstein2025avat3r}, we train on \num{244} subjects with \num{1000} frames each and evaluate on the remaining \num{12} subjects using approximately \num{1000} sampled frames.

\noindent \textbf{Expression Codes.} 
We generate the expression codes using the ARKit~\cite{arkit} face tracker, which fits a set of 51 blendshape bases to each frame, yielding per-frame weights $\mathbf{w} \in \mathbb{R}^{51}$. Each dimension corresponds to a semantically meaningful facial action (e.g. \texttt{left eye blink} or \texttt{jaw open}).

\subsection{Experimental Setup}

\textbf{Training Setup.} We follow a two-stage training strategy. 
First, we freeze the $H_E$ and $H_D$ (see Fig.~\ref{fig:architecture}) pretrained on Internal10K with $H_{UV} = W_{UV} = 256$ and train the remaining modules, $F_C$, $F_D$ and  $E$, with $N$ input frames and one target frame sampled from random viewpoints and expressions. In the second stage, we first finetune $H_E$ and $H_D$ and then the remaining modules on Ava-256. In this stage, we use furthest distance viewpoint sampling for input frame selection in both training and validation, following \cite{kirschstein2025avat3r, ji2026fastgha}, and a random target frame. We use HeadsUp~\cite{headsup} as a prior  $H_E$ and $H_D$.  The input frames are downsampled to 
$ 500\times375$ for Internal10K and to $512\times334$ for Ava-256. 
More training details in the Supp. Mat.~\ref{app:training_details}.

\noindent  \textbf{Metrics.}
We evaluate rendering quality using 
Peak Signal-to-Noise Ratio~(PSNR), Structural Similarity~(SSIM), and Learned Perceptual Image Patch Similarity~(LPIPS)~\cite{zhang2018unreasonable}. Following Avat3r~\cite{kirschstein2025avat3r}, we also report two face-specific metrics: Average Keypoint Distance~(AKD) which measures the distance in pixels between 2D keypoints estimated from PIPNet~\cite{jin2021pixel}, and cosine similarity~(CSIM) of ArcFace identity embeddings~\cite{deng2019arcface}. To measure body stability, we report $\sigma^t_\text{body}$, the mean per-pixel standard deviation of the body-view render (on the 0--255 RGB scale), computed across the different expressions of a subject. We report complementary perceptual and flow-based variants of this metric in Supp.~Mat.~\ref{app:temporal_body_metrics}. To measure cross-identity retargeting we compute Average Expression Distance (AED), calculated as the Euclidean distance between the reference and predicted expression codes extracted using ARKit \cite{arkit}.

\noindent  \textbf{Baselines.}
We compare against Avat3r~\cite{kirschstein2025avat3r}, a feed-forward method that fuses multi-view DUSt3R~\cite{wang2024dust3r} and Sapiens~\cite{khirodkar2024sapiens} features via cross-attention conditioned on expression tokens. We also compare against FastGHA~\cite{ji2026fastgha}, which predicts pixel-aligned Gaussians using a transformer conditioned on DINOv3~\cite{simeoni2025dinov3}  and Stable Diffusion VAE~\cite{rombach2022high} features with per-Gaussian deformation MLPs for reenactment. 
Due to the absence of publicly available code for both methods, they were implemented as closely as possible to the corresponding publications. 
We refer to Supp. Mat.~\ref{app:baselines_reimp}  and \ref{app:uika} for the reimplementation details and an additional comparison with UIKA~\cite{wu2026uika}.


\begin{table}[t]
\centering
  \vspace{1pt}
    \setlength{\tabcolsep}{3pt}
    \begin{tabular}{lcccccccc}
  \toprule
   & $\sigma^t_\text{body}$$\downarrow$ & \#G$\downarrow$
  & LPIPS$\downarrow$ & SSIM$\uparrow$ & AKD$\downarrow$
  & CSIM$\uparrow$ & PSNR$\uparrow$ & ms$\downarrow$ \\
  \midrule
  Avat3r  & 18.8 & 671K & 0.23 & 0.69 & 3.5 & 0.62 & 21.7 & 166 \\
  FastGHA & 22.9 & 534K & 0.23 & \underline{0.75} & 3.7 & 0.51 & 21.3 & 26.2 \\
  \midrule
  Ours$_\text{+BAN}$ & ~\textbf{0.7} & \textbf{65K} & \underline{0.21} & 0.74 & \textbf{3.1} & \textbf{0.67} & \underline{21.9} & \textbf{1.92} \\
  Ours    & ~\underline{6.0} & \textbf{65K} & \textbf{0.20} & \textbf{0.76} & \textbf{3.1} & \textbf{0.67} & \textbf{22.2} & \textbf{1.92} \\
  \bottomrule
  \end{tabular}%
  \vspace{5pt}
\caption{\textbf{Results on Ava-256 using 4 input views.} \#G is the number of Gaussians; \emph{$\sigma^t_\text{body}$} measures body stability, the mean per-pixel standard deviation of the body-view render across expressions (0--255 scale, lower is more stable); runtime in ms on A100 GPU. Ours and Ours$_\text{+BAN}$ achieve state-of-the-art results across  metrics.} 
  \vspace{-20pt}
\label{tab:Viewsexperiments}
\end{table}


\subsection{Main Results}

We report Ours$_\text{+BAN}$, supervised on body-aligned targets (Sec.~\ref{sec:body-alignment}), and Ours, supervised against raw captures, matching the supervision regime of prior feed-forward methods, to enable direct pixel-metric comparison with baselines.

\noindent  \textbf{Quantitative Results.}
Tab.~\ref{tab:Viewsexperiments} and Tab.~\ref{tab:merged_quality_runtime} report metrics on Ava-256 and Internal10K. Our method consistently achieves the best AKD and CSIM scores on both datasets across all input counts, indicating superior expression driving and identity preservation while using $8\times$ fewer Gaussians and maintaining constant runtime. LPIPS and SSIM remain close to the best baseline across datasets. 
Ours$_\text{+BAN}$ also has the most temporally stable body, with $\sigma^t_\text{body}$ more than an order of magnitude below the baselines (Tab.~\ref{tab:Viewsexperiments}).

\noindent  \textbf{Qualitative Results.} 
Fig.~\ref{fig:4_ava256_renders} shows visual comparisons on Ava-256. Our method produces sharper facial details and more accurate expression transfer. FastGHA tends to produce over-smoothed textures and Avat3r does not preserve the identity well. Fig. \ref{fig:view_scaling} in Supp. Mat. compares all methods on Internal10K for different numbers of input images. Our method shows better identity preservation and passthrough details with high fidelity. We further illustrate the improvements of BAN in  Fig.~\ref{fig:body_alignment_vis} and Fig.~\ref{fig:ava256_cross_expr_viz}, where Ours$_\text{+BAN}$ produces a sharper body and eliminates jitter. The Supp. Mat. provides videos and an in-the-wild comparison (Supp. Mat.~\ref{sec:suppITW}).

\begin{table}[t]
\centering
\setlength{\tabcolsep}{3pt}
\resizebox{\columnwidth}{!}{%
\begin{tabular}{lrccccccrrrr}
  \toprule
    & & & & & & & & \multicolumn{2}{c}{A100} & \multicolumn{2}{c}{iPhone 17 Pro} \\
    \cmidrule(lr){9-10} \cmidrule(lr){11-12} & \#I & \#G & LPIPS$\downarrow$ & SSIM$\uparrow$ & AKD$\downarrow$ & CSIM$\uparrow$ & PSNR$\uparrow$
    & ms$\downarrow$ & FPS$\uparrow$ & ms$\downarrow$ & FPS$\uparrow$ \\
  \midrule
  Avat3r  & 1 & 182K         & \textbf{0.26}   & 0.73            & \underline{9.1} & 0.44             & \underline{20.5} & 43.20         & 23           & --            & --           \\
  FastGHA & 1 & 150K         & \underline{0.27}& \textbf{0.76}   & 9.6             & \underline{0.56} & 20.3             & 6.81          & 147          & 11.80         & 85           \\
  Ours    & 1 & \textbf{65K} & \underline{0.27}& \textbf{0.76}   & \textbf{7.9}    & \textbf{0.62}    & \textbf{20.8}    & \textbf{1.92} & \textbf{522} & \textbf{1.45} & \textbf{690} \\
  \midrule
  Avat3r  & 2 & 365K         & \textbf{0.26}   & 0.74            & \underline{8.8} & 0.56             & \underline{20.8} & 83.79         & 12           & --            & --           \\
  FastGHA & 2 & 301K         & 0.27            & \textbf{0.77}   & 9.3             & \underline{0.58} & 20.7             & 12.97         & 77           & 24.36         & 41           \\
  Ours    & 2 & \textbf{65K} & \textbf{0.26}   & \underline{0.76}& \textbf{7.4}    & \textbf{0.65}    & \textbf{21.3}    & \textbf{1.92} & \textbf{522} & \textbf{1.45} & \textbf{690} \\
  \midrule
  Avat3r  & 4 & 730K         & \textbf{0.24}   & 0.74            & 8.3             & 0.60             & \underline{21.1} & 166.23        & 6            & --            & --           \\
  FastGHA & 4 & 602K         & 0.26            & \textbf{0.77}   & 9.3             & 0.58             & 20.9             & 26.25         & 38           & 45.14         & 22           \\
  Ours$_\text{+BAN}$ & 4 & \textbf{65K} & 0.28 & 0.73            & \underline{7.5} & \underline{0.66} & 19.8             & \textbf{1.92} & \textbf{522} & \textbf{1.45} & \textbf{690} \\
  Ours    & 4 & \textbf{65K} & \textbf{0.24}   & \textbf{0.77}   & \textbf{7.4}    & \textbf{0.67}    & \textbf{21.6}    & \textbf{1.92} & \textbf{522} & \textbf{1.45} & \textbf{690} \\
  \bottomrule
\end{tabular}%
}
\vspace{4pt}
\caption{\textbf{Quality and runtime on Internal10K across varying numbers of input images (\#I).} \#G is the number of Gaussians. Runtime is the animation-stage time; Ours is up to $13\times$ faster than FastGHA on the A100 and $31\times$ on iPhone 17 Pro Neural Engine (ANE). Pixel-aligned metrics for Ours$_\text{+BAN}$ degrade because the predicted body is aligned with a reference image instead of the ground truth. This does not affect animation (AKD) and identity preservation (CSIM) quality.}
  \vspace{-20pt}
\label{tab:merged_quality_runtime}
\end{table}


\subsection{Analysis and Ablation Study} \label{subsec:experiments-ablation}

\begin{table}[p]
  \centering
  \setlength{\tabcolsep}{5pt}
  \begin{tabular}{lcccc}
  \toprule
  \tikz[baseline=(k.base)]\node[draw=black!45, dashed, inner sep=1.5pt, rounded corners=1pt](k){\footnotesize Ours $\to$ Ours$_\text{+BAN}$}; & \multicolumn{2}{c}{Face Region} & \multicolumn{2}{c}{Body Region} \\
  \cmidrule(lr){2-3} \cmidrule(lr){4-5}
  Dataset & SSIM$\uparrow$ & PSNR$\uparrow$ & SSIM$\uparrow$ & PSNR$\uparrow$ \\
  \midrule
  Ava-256     & 0.75 $\to$ 0.75 & 23.4 $\to$ 23.4 & 0.87 $\to$ 0.83 & 24.5 $\to$ 23.8 \\
  Internal10K & 0.87 $\to$ 0.87 & 28.4 $\to$ 28.1 & 0.84 $\to$ 0.80 & 24.5 $\to$ 23.9 \\
  \bottomrule
  \end{tabular}
 \vspace{4pt}
\caption{
\textbf{Regional image quality (Ava-256, Internal10K)}. We report SSIM and PSNR on GT mask-segmented face and body regions at a fixed cropped frontal camera view. Note that these regional metrics do not directly compare to Tab.~\ref{tab:Viewsexperiments} or Tab.~\ref{tab:merged_quality_runtime}.
}
\vspace{-20pt}
  \label{tab:ban_regional}
  \end{table}

\textbf{Ablation over the number of input images.} As shown in Tab.~\ref{tab:merged_quality_runtime}, increasing the number of input views improves quality for all methods. Our method achieves superior AKD and CSIM at every view count, with comparable SSIM and LPIPS, while using up to 8$\times$ fewer Gaussians than the baselines and constant runtime. Qualitative comparison can be found in the Supp. Mat.~\ref{sec:suppNViews}.

\noindent  \textbf{Canonical Representation.}
We ablate the learned canonical representation by replacing $\mathbf{C}$ in Eq.~(\ref{eq:residual}) with the Gaussian UV map corresponding to one of the input images. This means that $E(\mathbf{w}, \mathbf{D})$ must account for the full range of expression variations. As shown in Tab.~\ref{tab:design_ablations} and Fig.~\ref{fig:design_ablations_qualitative}(c), this leads to worse metrics, as $E$ must learn larger residuals to animate an arbitrary $\mathbf{G}_1$ and correct identity inconsistencies.
  


\begin{figure*}[p]
  \centering
  \begin{tikzpicture}
    \node[anchor=south west, inner sep=0] (img)
      {\includegraphics[width=\linewidth]{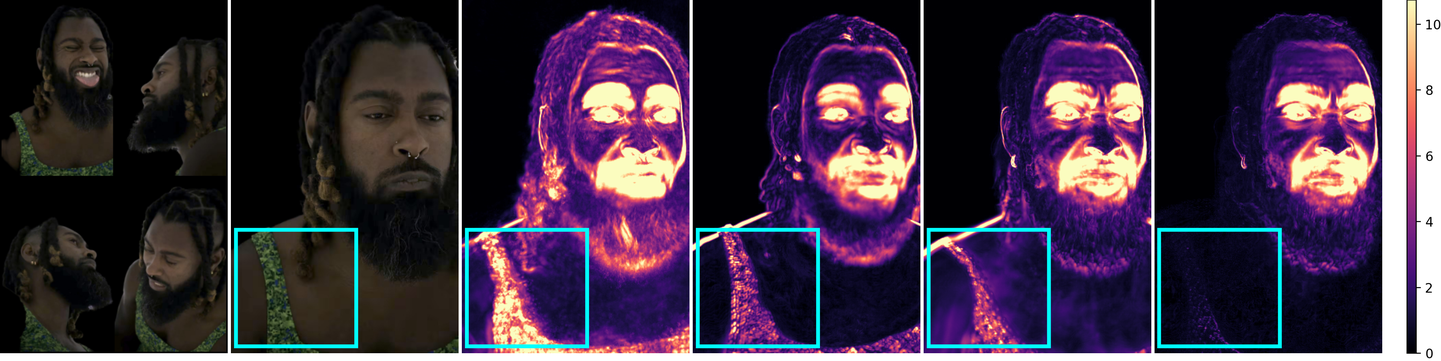}};

    \begin{scope}[
      every node/.style={anchor=base, font=\normalsize, yshift=2pt}
    ]
      \node at ([xshift=0.080\linewidth]img.north west) {Input};
      \node at ([xshift=0.238\linewidth]img.north west) {Sample Frame};
      \node at ([xshift=0.397\linewidth]img.north west) {Avat3r};
      \node at ([xshift=0.556\linewidth]img.north west) {FastGHA};
      \node at ([xshift=0.715\linewidth]img.north west) {Ours};
      \node at ([xshift=0.875\linewidth]img.north west) {Ours$_\text{+BAN}$};
    \end{scope}
  \end{tikzpicture}
  \vspace{-1em}
  \caption{\textbf{Temporal stability.} 
   The colors show the per-pixel temporal standard deviation across a sequence driven by an expression code
   See
   zoomed cyan boxes
   for
   jitter unrelated to the expression. Avat3r~\cite{kirschstein2025avat3r},
   FastGHA~\cite{ji2026fastgha}, and Ours all exhibit this jitter, whereas
   Ours$_\text{+BAN}$, trained against body-aligned supervision targets (Fig.~\ref{fig:gt_body_align_architecture}), keeps the body stable. The full
   video is provided in the Supp. Mat.}
  \vspace{-10pt}
  \label{fig:body_alignment_vis}
\end{figure*}

\input{figures/ava256_crossexpr_comp}


\noindent \textbf{Deformation Latents.} We remove the deformation latents and feed only the expression code to the animation network in Eq.~(\ref{eq:residual}). As shown in Tab.~\ref{tab:design_ablations} and Fig.~\ref{fig:design_ablations_qualitative}(e), this leads to loss of subject-specific details and worse identity preservation across expressions, as the expression codes are subject-agnostic.


\newcommand{\imgzoomcompBR}[7]{%
\begin{tikzpicture}[inner sep=0pt, outer sep=0pt]
  \node[anchor=south west] (img) at (0.0,0.0)
    {#1};
  \begin{scope}[x={(img.south east)}, y={(img.north west)}]
    \draw[cyan, line width=0.2pt] (#2,#3) rectangle (#4,#5);
    \pgfmathsetmacro{\cropw}{#4-#2}
    \pgfmathsetmacro{\croph}{#5-#3}
    \pgfmathsetmacro{\zoomh}{#6}
    \pgfmathsetmacro{\zoomw}{\zoomh*\cropw/\croph}
    \pgfmathsetmacro{\zoomscale}{\zoomh/\croph}
    \begin{scope}
      \clip ({0.98-\zoomw},0.02) rectangle (0.98,{0.02+\zoomh});
      \node[anchor=south west, inner sep=0pt, scale=\zoomscale]
        at ({0.98-\zoomw-#2*\zoomscale},{0.02-#3*\zoomscale})
        {#1};
    \end{scope}
    \draw[cyan, line width=0.5pt] ({0.98-\zoomw},0.02) rectangle (0.98,{0.02+\zoomh});
  \end{scope}
\end{tikzpicture}%
}

\newcommand{\imgzoomcompTL}[7]{%
\begin{tikzpicture}[inner sep=0pt, outer sep=0pt]
  \node[anchor=south west] (img) at (0.0,0.0)
    {#1};
  \begin{scope}[x={(img.south east)}, y={(img.north west)}]
    \draw[cyan, line width=0.2pt] (#2,#3) rectangle (#4,#5);
    \pgfmathsetmacro{\cropw}{#4-#2}
    \pgfmathsetmacro{\croph}{#5-#3}
    \pgfmathsetmacro{\zoomh}{#6}
    \pgfmathsetmacro{\zoomw}{\zoomh*\cropw/\croph}
    \pgfmathsetmacro{\zoomscale}{\zoomh/\croph}
    \begin{scope}
      \clip (0.02,{0.98-\zoomh}) rectangle ({0.02+\zoomw},0.98);
      \node[anchor=south west, inner sep=0pt, scale=\zoomscale]
        at ({0.02-#2*\zoomscale},{0.98-\zoomh-#3*\zoomscale})
        {#1};
    \end{scope}
    \draw[cyan, line width=0.5pt] (0.02,{0.98-\zoomh}) rectangle ({0.02+\zoomw},0.98);
  \end{scope}
\end{tikzpicture}%
}

\begin{figure*}[t]
\centering
\setlength{\tabcolsep}{0pt}
\renewcommand{\arraystretch}{0}
\footnotesize
\scalebox{0.95}{
\begin{tabular}{@{}cccccc@{}}
Input & Avat3r & FastGHA  & Ours & Ours$_\text{+BAN}$ & GT \\[4pt]


\includegraphics[height=0.19\linewidth, trim=0 0 13 0]{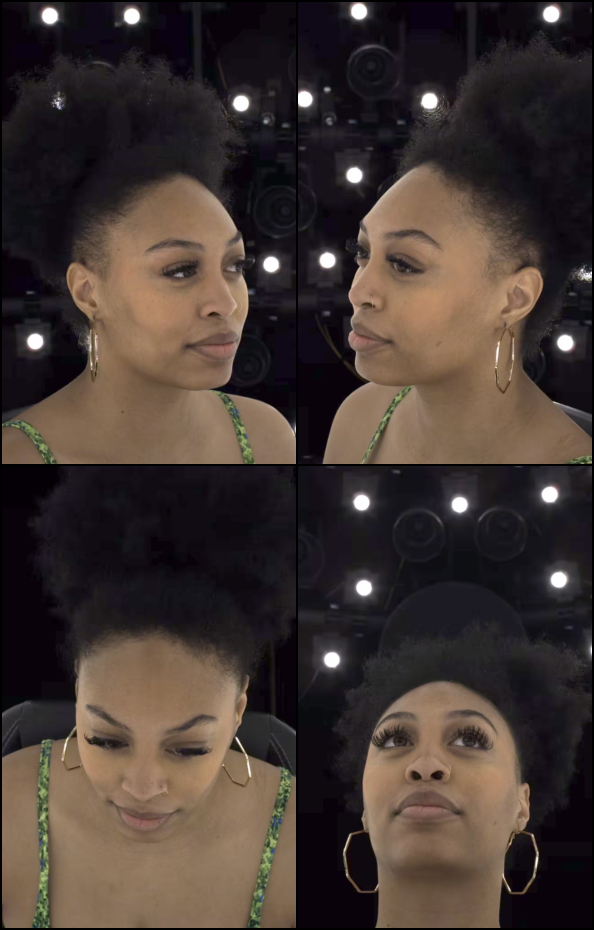} &
\imgzoomcompTL{\includegraphics[height=0.19\linewidth, trim=0 20 0 110, clip]{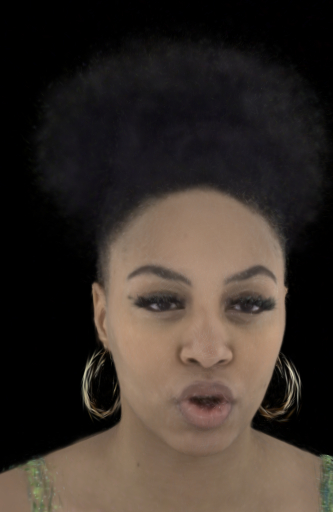}}{0.50}{0.12}{0.73}{0.33}{0.40}{0 20 0 110} &
\imgzoomcompTL{\includegraphics[height=0.19\linewidth, trim=0 20 0 110, clip]{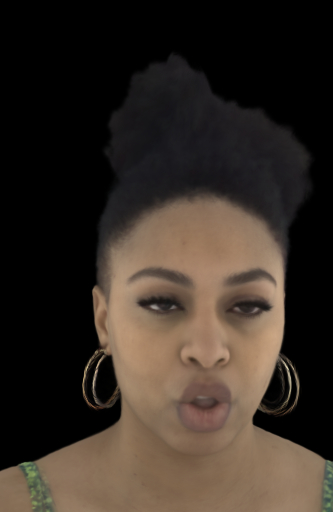}}{0.50}{0.12}{0.73}{0.33}{0.40}{0 20 0 110} &
\imgzoomcompTL{\includegraphics[height=0.19\linewidth, trim=0 20 0 110, clip]{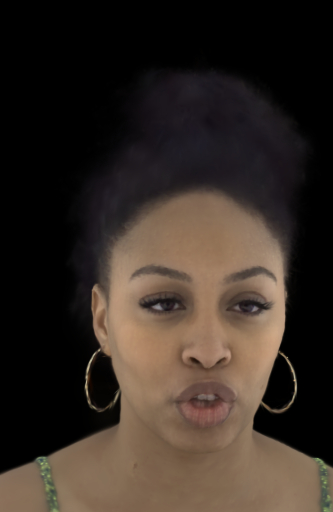}}{0.50}{0.12}{0.73}{0.33}{0.40}{0 20 0 110} &
\imgzoomcompTL{\includegraphics[height=0.19\linewidth, trim=0 20 0 110, clip]{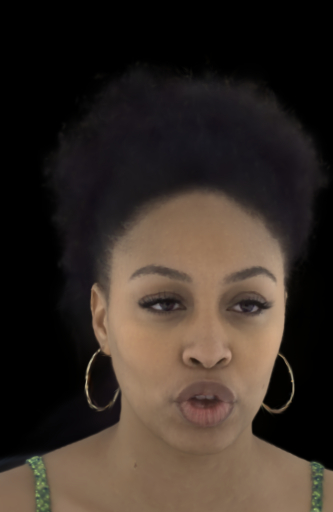}}{0.50}{0.12}{0.73}{0.33}{0.40}{0 20 0 110} &
\imgzoomcompTL{\includegraphics[height=0.19\linewidth, trim=0 20 0 110, clip]{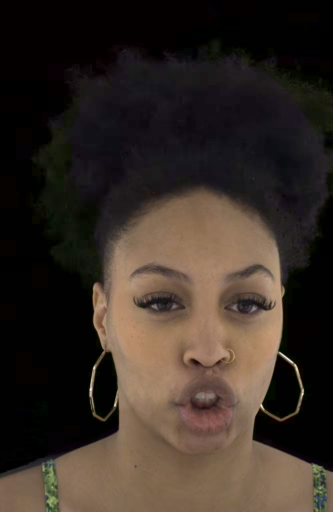}}{0.50}{0.12}{0.73}{0.33}{0.40}{0 20 0 110} \\

\includegraphics[height=0.19\linewidth]{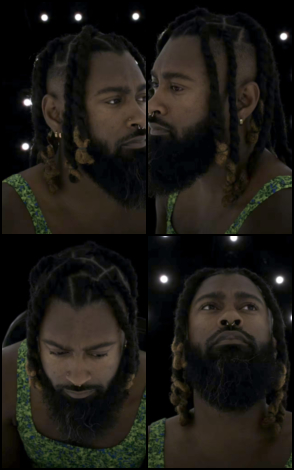} &
\imgzoomcompTL{\includegraphics[height=0.19\linewidth, trim=0 120 0 10, clip]{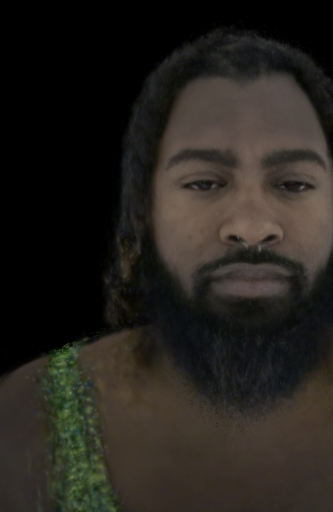}}{0.05}{0.01}{0.30}{0.20}{0.30}{0 120 0 10} &
\imgzoomcompTL{\includegraphics[height=0.19\linewidth, trim=0 120 0 10, clip]{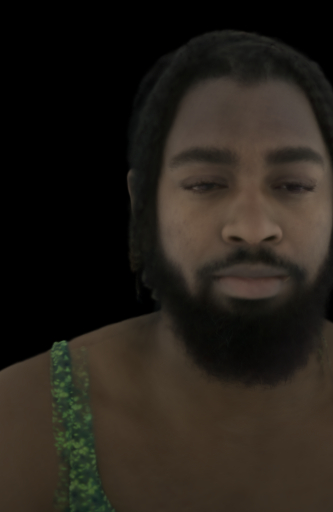}}{0.05}{0.01}{0.30}{0.20}{0.30}{0 120 0 10} &
\imgzoomcompTL{\includegraphics[height=0.19\linewidth, trim=0 120 0 10, clip]{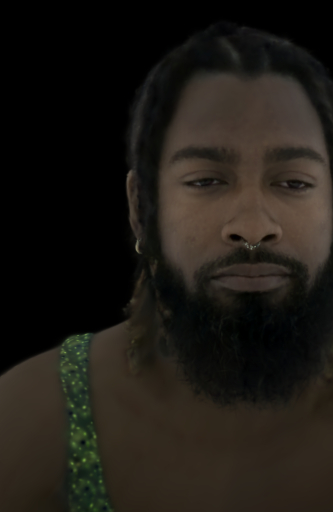}}{0.05}{0.01}{0.30}{0.20}{0.30}{0 120 0 10} &
\imgzoomcompTL{\includegraphics[height=0.19\linewidth, trim=0 120 0 10, clip]{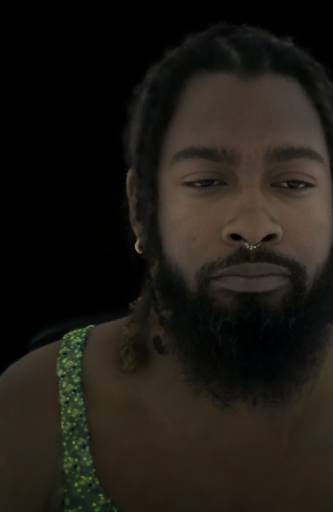}}{0.05}{0.01}{0.30}{0.20}{0.30}{0 120 0 10} &
\imgzoomcompTL{\includegraphics[height=0.19\linewidth, trim=0 120 0 10, clip]{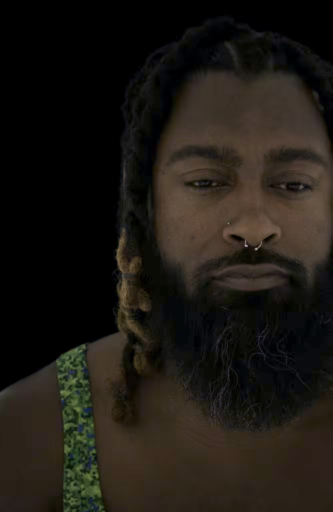}}{0.05}{0.01}{0.30}{0.20}{0.30}{0 120 0 10} \\
\end{tabular}%
}
 \vspace{-5pt}
\caption{\textbf{Qualitative comparison on Ava-256.} Our method produces sharper facial details (e.g. eyes/teeth/wrinkles) and better expression transfer. FastGHA \cite{ji2026fastgha} tends to produce over-smoothed textures and Avat3r~\cite{kirschstein2025avat3r} does not preserve identity as well. The boxes mark the zoom-in region shown as an inset in each panel.
}

\label{fig:4_ava256_renders}
\vspace{-10pt}
\end{figure*}

\noindent \textbf{Body Alignment Network.}
 BAN composites supervision targets that combine the reference body pose with the target facial expression (Fig.~\ref{fig:gt_body_align_architecture}). It removes residual pose misalignments from the supervisory signal. Training against these targets eliminates expression-independent body jitter at inference time, reducing $\sigma^t_\text{body}$ by nearly an order of magnitude (Fig.~\ref{fig:body_alignment_vis}, Tab.~\ref{tab:Viewsexperiments}), and produces a sharper body (Fig.~\ref{fig:ava256_cross_expr_viz}). This does not degrade the face: face-specific metrics (AKD, CSIM) remain superior (Tab.~\ref{tab:Viewsexperiments}, Tab.~\ref{tab:merged_quality_runtime}). The full-image pixel-aligned metrics (LPIPS, SSIM, PSNR) drop slightly because the body is aligned to a reference instead of the ground truth, not because face quality is compromised. A regional breakdown (Tab.~\ref{tab:ban_regional}) shows that only the body-region scores change, while the face-region quality is essentially unchanged, so identity and animation quality are preserved. The drop is larger on Internal10K (Tab.~\ref{tab:merged_quality_runtime}) than on Ava-256 (Tab.~\ref{tab:Viewsexperiments}). Its validation set has more extreme viewpoints, where the body fills more of the frame and the body-alignment cost is larger than at the frontal camera.

  

\noindent  \textbf{Passthrough Loss.}
As shown in Tab.~\ref{tab:design_ablations} and Fig.~\ref{fig:design_ablations_qualitative}(d), the passthrough auxiliary loss further improves results by encouraging the network to preserve texture details and enforce geometric consistency with the input images.

\begin{table}[t]
\centering
    \setlength{\tabcolsep}{2pt}
    \begin{tabular}{lccccc}
  \toprule
  \multicolumn{6}{c}{\textbf{Internal10K Dataset}} \\
  Configuration & PSNR$\uparrow$ & SSIM$\uparrow$ & LPIPS$\downarrow$ & AKD$\downarrow$ & CSIM$\uparrow$ \\
  \midrule
  Full model (Ours) & \textbf{21.6} & \textbf{0.77} & \textbf{0.24} & \textbf{7.4} & \textbf{0.67} \\
  w/o canonical representation & 21.2 & 0.76 & 0.27 & 7.5 & 0.65 \\
  w/o passthrough loss & 21.3 & \textbf{0.77} & 0.25 & 7.5 & 0.66 \\
  w/o deformation latent & 21.3 & 0.75 & 0.25 & 9.3 & 0.59 \\
  Full model + BAN & 19.8 & 0.73 & 0.28 & 7.5 & 0.66 \\
  \bottomrule
  \end{tabular}%
  \vspace{4pt}
\caption{\textbf{Design choices ablation.} We individually ablate key components of our pipeline. Each \emph{w/o} row disables one component while keeping all others at default.}
\vspace{-25pt}
\label{tab:design_ablations}
\end{table}

\newcommand{\imgzoomablation}[6]{%
\begin{tikzpicture}[inner sep=0pt, outer sep=0pt]
  \node[anchor=south west] (img) at (0.0,0.0) {#1};
  \begin{scope}[x={(img.south east)}, y={(img.north west)}]
    \draw[cyan, line width=0.2pt] (#2,#3) rectangle (#4,#5);
    \pgfmathsetmacro{\cropw}{#4-#2}
    \pgfmathsetmacro{\croph}{#5-#3}
    \pgfmathsetmacro{\zoomh}{#6}
    \pgfmathsetmacro{\zoomw}{\zoomh*\cropw/\croph}
    \pgfmathsetmacro{\zoomscale}{\zoomh/\croph}
    \begin{scope}
      \clip (0.02,0.02) rectangle ({0.02+\zoomw},{0.02+\zoomh});
      \node[anchor=south west, inner sep=0pt, scale=\zoomscale]
        at ({0.02-#2*\zoomscale},{0.02-#3*\zoomscale})
        {#1};
    \end{scope}
    \draw[cyan, line width=0.5pt] (0.02,0.02) rectangle ({0.02+\zoomw},{0.02+\zoomh});
  \end{scope}
\end{tikzpicture}%
}

\begin{figure*}[ht!]
\centering
\setlength{\tabcolsep}{1pt} 
\renewcommand{\arraystretch}{0.5}
\scriptsize

\scalebox{0.9}{
\begin{tabular}{@{}ccccc@{}}
(a) GT & (b) Full & (c) w/o Can. & (d) w/o Pass. & (e) w/o Def. \\


\imgzoomablation{\includegraphics[width=0.15\linewidth]{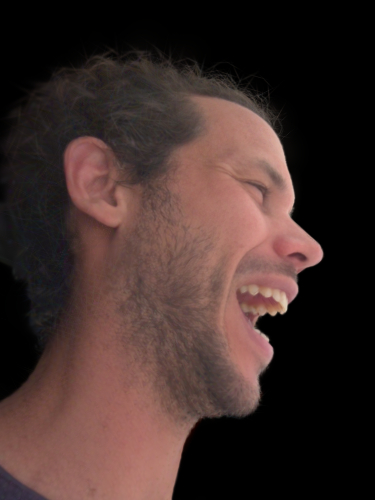}}{0.60}{0.26}{0.81}{0.47}{0.40} &
\imgzoomablation{\includegraphics[width=0.15\linewidth]{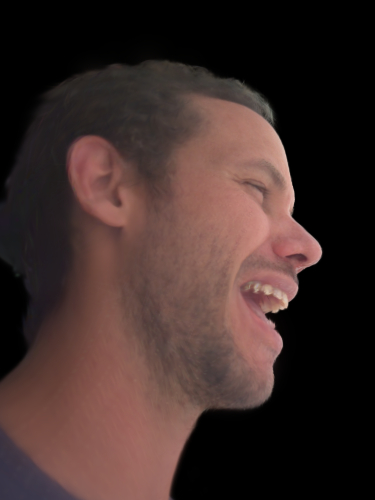}}{0.60}{0.26}{0.81}{0.47}{0.40} &
\imgzoomablation{\includegraphics[width=0.15\linewidth]{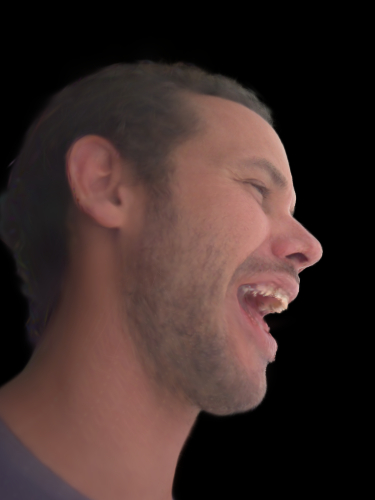}}{0.60}{0.26}{0.81}{0.47}{0.40} &
\imgzoomablation{\includegraphics[width=0.15\linewidth]{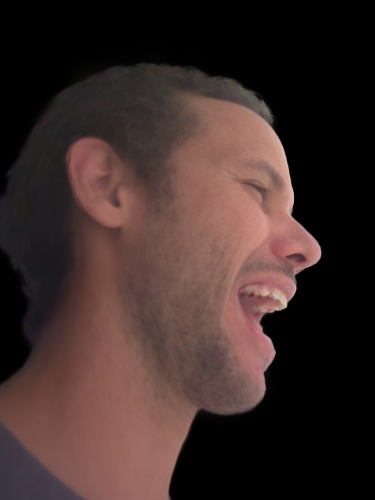}}{0.60}{0.26}{0.81}{0.47}{0.40} &
\imgzoomablation{\includegraphics[width=0.15\linewidth]{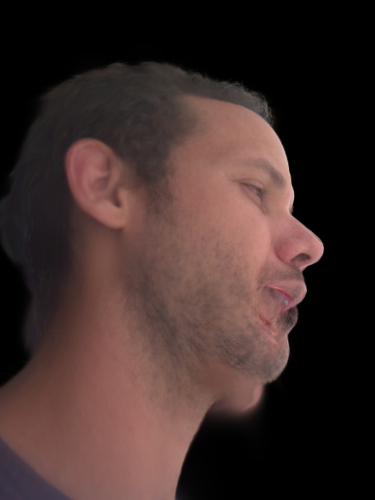}}{0.60}{0.26}{0.81}{0.47}{0.40} \\

\imgzoomablation{\includegraphics[width=0.15\linewidth]{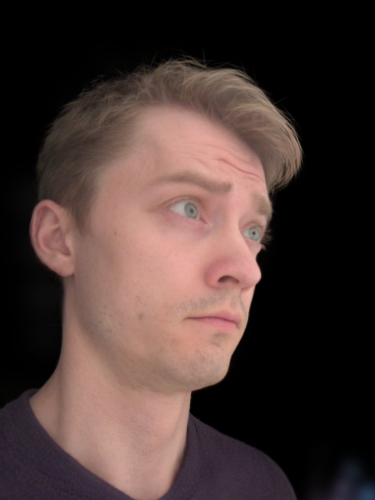}}{0.45}{0.30}{0.73}{0.65}{0.45} &
\imgzoomablation{\includegraphics[width=0.15\linewidth]{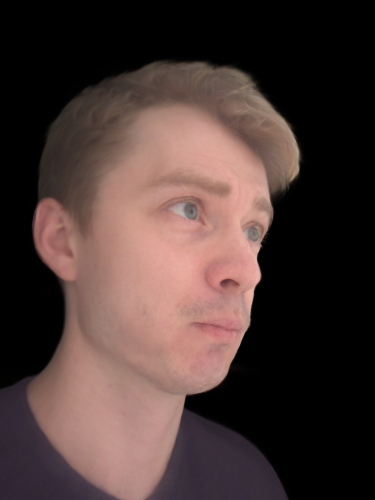}}{0.45}{0.30}{0.73}{0.65}{0.45} &
\imgzoomablation{\includegraphics[width=0.15\linewidth]{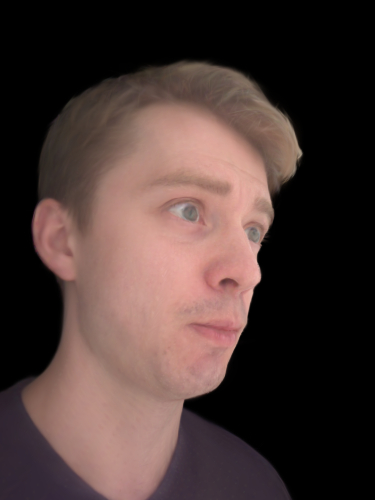}}{0.45}{0.30}{0.73}{0.65}{0.45} &
\imgzoomablation{\includegraphics[width=0.15\linewidth]{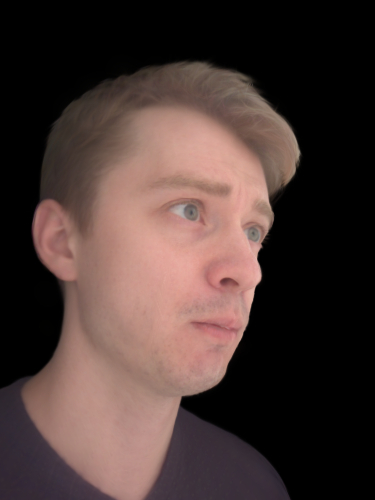}}{0.45}{0.30}{0.73}{0.65}{0.45} &
\imgzoomablation{\includegraphics[width=0.15\linewidth]{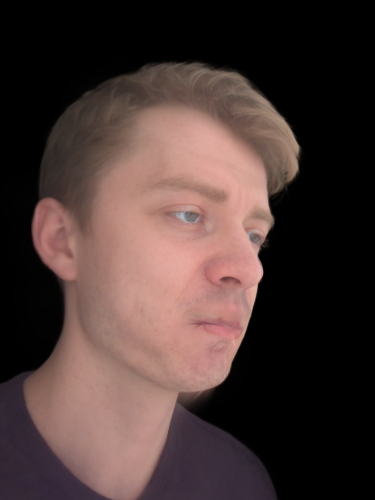}}{0.45}{0.30}{0.73}{0.65}{0.45} \\

\end{tabular}
}
\vspace{-4pt}
\caption{\textbf{Qualitative evaluation of our design choices.} We compare (a) the ground truth (GT) against (b) our full model and ablations removing (c) the canonical representation, (d) the passthrough loss, and (e) the deformation latents. Cyan boxes mark the zoom-in region shown in the bottom-left of each panel.}
\vspace{-4pt}
\label{fig:design_ablations_qualitative}
\vspace{-5pt}
\end{figure*}



\begin{figure*}[t]
  \centering
  
    \begin{minipage}[c]{0.7\textwidth}
      \centering
      \begin{tikzpicture}
        \node[anchor=south west, inner sep=0] (img)
          {\includegraphics[width=\linewidth]{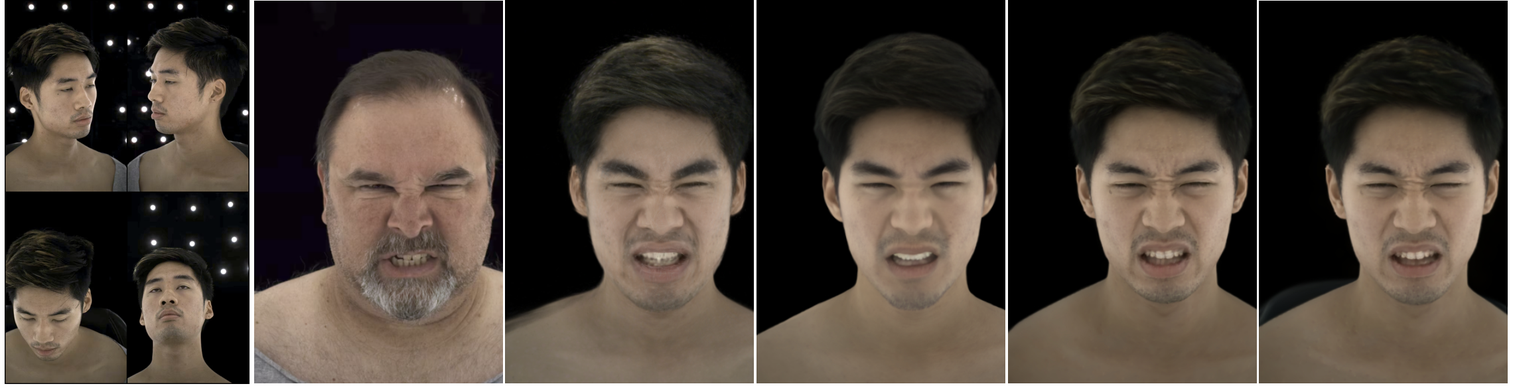}};
        \begin{scope}[every node/.style={anchor=base, font=\scriptsize, yshift=2pt}]
          \node at ([xshift=0.084\linewidth]img.north west) {Input};
          \node at ([xshift=0.250\linewidth]img.north west) {Expression};
          \node at ([xshift=0.416\linewidth]img.north west) {Avat3r};
          \node at ([xshift=0.582\linewidth]img.north west) {FastGHA};
          \node at ([xshift=0.748\linewidth]img.north west) {Ours};
          \node at ([xshift=0.914\linewidth]img.north west) {Ours$_\text{+BAN}$};
        \end{scope}
      \end{tikzpicture}
    \end{minipage}\hfill
  \begin{minipage}[c]{0.27\textwidth}
    \centering
    \vspace{1pt}
    \setlength{\tabcolsep}{2pt}
    \scriptsize
    \begin{tabular}{lc}
      \toprule
      Method & AED$\times 10^{-2} \downarrow$ \\
      \midrule
      Avat3r  & 5.97 \\
      FastGHA & 6.00 \\
      \midrule
      Ours    & 5.43 \\
      Ours$_\text{+BAN}$ & \textbf{5.35} \\
      \bottomrule
    \end{tabular}
    \vspace{1pt}
  \end{minipage}
  
  \vspace{-5pt}
  \caption{
    \textbf{Cross-identity retargeting.} For each pair, the identity input is shown in the first column, and the expression is driven by the subject in the second column. \emph{Right table:} Average Expression Distance (AED) metric.
  }
  \vspace{-15pt}
  \label{fig:ava256_cross_identity}
\end{figure*}

\noindent  \textbf{Cross-Identity Retargeting.} 
In Fig.~\ref{fig:ava256_cross_identity}, we demonstrate the cross-identity retargeting. Avat3r shows artifacts around the shoulders while FastGHA loses fine details like wrinkles. Our method  transfers the expressions while keeping sharpness and details. The table on the right quantitatively confirms that our method better preserves expressivity, as measured by AED.



\noindent  \textbf{Runtime Performance.} We benchmark the inference time 
on an NVIDIA A100 GPU and iPhone~17 Pro (ANE), reporting only the animation runtime for all methods, after precomputing all expression-independent parts (\ie, the canonical representation for FastGHA and ours). 
As shown in Tab.~\ref{tab:merged_quality_runtime} and 
Fig.~\ref{fig:teaser} (right), our method achieves constant latency regardless of the number of input frames, while FastGHA scales linearly. 
We run 3$\times$ to 31$\times$ faster with up to 8$\times$ fewer Gaussians while improving animation quality across evaluated setups (A100 to iPhone 17 Pro ANE).
Including rendering, our end-to-end latency on an A100 is 3.25\,ms, 8.5$\times$ faster than FastGHA and 52$\times$ faster than Avat3r (Supp. Tab.~\ref{tab:runtime_e2e}). Supp. Mat.~\ref{sec:suppRuntime} and \ref{sec:suppIdentityNetworkPerformance} report further runtime analyses.




\vspace{5pt}
\section{Conclusion}
We presented GHARP, a method for real-time animation of 3D Gaussian head avatars on top of a learned canonical Gaussian representation. This formulation allows us to capture the subject appearance and geometry in a fixed-size set of Gaussians, thus decoupling animation cost from the number of input views. Additionally, a body alignment network (BAN) resolves appearance inconsistencies in the body region during training, improving temporal stability at animation time. Experiments demonstrate state-of-the-art quality while animating up to \num{8}x fewer Gaussians with an order of magnitude lower inference cost.
Limitations of the proposed method include the dependence on the calibrated camera poses; therefore, inaccurate estimates could propagate through the encoder.
Future extensions could include robustness to temporal variations or noise in the input signal and material factorization for portrait relighting.


\bibliographystyle{splncs04}
\bibliography{main}


\newpage

\appendix
  \section{Supplementary Material}
\FloatBarrier
\subsection{Network Architecture}
\label{app:architecture_details}




\textbf{Canonical Fusion Network $F_C$.} $F_C$ is a fully convolutional residual network operating at $32{\times}32$ spatial resolution. It receives the concatenation of per-view latents $\{\mathbf{Z}_i = H_E(I_i)\}_{i=1}^{N}$, each $\mathbf{Z}_i \in \mathbb{R}^{32 \times 32 \times 512}$ together with the concatenation of their respective 6D Plücker embeddings, along the channel dimension, yielding an input $\in \mathbb{R}^{32 \times 32 \times 2072}$ (for $N{=}4$ views). The output $\in \mathbb{R}^{32 \times 32 \times 2560}$ is split into $N$ spatial blending weight maps $\{b_i\}_{i=1}^{N} \in \mathbb{R}^{32 \times 32 \times 512}$ and a correction term $\boldsymbol{\delta}_c \in \mathbb{R}^{32 \times 32 \times 512}$. The last convolution is zero-initialized so the network initially outputs the unweighted mean of the input latents. We describe the full architecture in Tab.~\ref{tab:arch_fc}.

\begin{table}[h]
\centering
\begin{tabular}{lc}
\toprule
Type & Channels \\
\midrule
Input (concat latents)                    & $N{\times}(512+6) = 2072$ \\
Conv $3{\times}3$                         & $2072 \to 512$ \\
ResNetBlock $\times 8$                    & $512 \to 512$ \\
GN16$\to$SiLU$\to$Conv $1{\times}1$      & $512 \to 2560$ \\
\bottomrule
\multicolumn{2}{l}{\footnotesize ResNetBlock = GN16$\to$SiLU$\to$Conv $3{\times}3$, repeated twice, with skip connection.} \\
\multicolumn{2}{l}{\footnotesize GN = Group Normalization layer with 16 groups} \\
\bottomrule
\end{tabular}
\vspace{5pt}
\caption{\textbf{Architecture of $F_C$} (canonical fusion network). Resolution is constant at $32{\times}32$.}
\label{tab:arch_fc}

\end{table}

\noindent \textbf{Deformation Encoder $F_D$.}
$F_D$ consists of two stages. First, a U-Net with shared weights processes each Gaussian UV map $\mathbf{G}_i$ for $i = 1, \ldots, N$ and the canonical $\mathbf{C}$ independently, each $\in \mathbb{R}^{256 \times 256 \times d_G}$ with $d_G{=}23$, producing per-map feature maps $\in \mathbb{R}^{64 \times 64 \times 192}$. Second, the $(N{+}1)$ feature maps are concatenated along the channel dimension into a tensor $\in \mathbb{R}^{64 \times 64 \times (N{+}1){\cdot}192}$ and consolidated by a residual network into the deformation latent $\mathbf{D} \in \mathbb{R}^{64 \times 64 \times 96}$.  We describe the full architecture in Tab.~\ref{tab:arch_fd_v1} and ~\ref{tab:arch_fd_v2}.

\begin{table}[h!]
\centering

\begin{subtable}[t]{\columnwidth}
\centering
\begin{tabular}{lcc}
\toprule
Type & Channels & Resolution \\
\midrule
\multicolumn{3}{l}{\textit{U-Net Encoder}} \\
Input ($\mathbf{G}_i$ or $\mathbf{C}$) & $d_G = 23$ & $256^2$ \\
Conv $2{\times}2$ s2, BN, ReLU & $23 \to 64$     & $256^2 \to 128^2$ \\
Conv $3{\times}3$ s2           & $64 \to 128$    & $128^2 \to 64^2$ \\
DownBlock                      & $128 \to 128$   & $64^2 \to 32^2$ \\
DownBlock                      & $128 \to 256$   & $32^2 \to 16^2$ \\
DownBlock                      & $256 \to 256$   & $16^2 \to 8^2$ \\
DownBlock                      & $256 \to 256$   & $8^2 \to 4^2$ \\
DownBlock                      & $256 \to 256$   & $4^2 \to 2^2$ \\
\midrule
\multicolumn{3}{l}{\textit{U-Net Decoder}} \\
UpBlock                        & $256 \to 256$   & $2^2 \to 4^2$ \\
UpBlock $+$ skip               & $512 \to 256$   & $4^2 \to 8^2$ \\
UpBlock $+$ skip               & $512 \to 256$   & $8^2 \to 16^2$ \\
UpBlock $+$ skip               & $512 \to 128$   & $16^2 \to 32^2$ \\
UpBlock $+$ skip               & $256 \to 64$    & $32^2 \to 64^2$ \\
\midrule
Concat skip                    & $64{+}128 \to 192$ & $64^2$ \\
\bottomrule
\multicolumn{3}{l}{\footnotesize DownBlock = Conv$\to$BN$\to$ReLU$\to$AvgPool$_2$;} \\
\multicolumn{3}{l}{\footnotesize UpBlock = Upsample$_2$$\to$Conv $3{\times}3$$\to$BN$\to$ReLU; $+$ = skip concatenation.} \\
\bottomrule
\end{tabular}

\end{subtable}
\vspace{5pt}
\caption{\textbf{Architecture of $F_D$} (deformation encoder). Per-Gaussian U-Net applied independently to each $\mathbf{G}_i$ and $\mathbf{C}$. }
\label{tab:arch_fd_v1}
\vspace{-15pt}
\end{table}

\begin{table}[h]
\centering

\begin{subtable}[t]{\columnwidth}
\centering
\begin{tabular}{lc}
\toprule
Type & Channels \\
\midrule
Input (concat all maps)              & $(N{+}1){\times}192 = 960$ \\
Conv $3{\times}3$                    & $960 \to 512$ \\
ResNetBlock $\times 2$               & $512 \to 512$ \\
GN16$\to$SiLU$\to$Conv $1{\times}1$ & $512 \to 96$ \\
\bottomrule
\multicolumn{2}{l}{\footnotesize ResNetBlock = GN16$\to$SiLU$\to$Conv $3{\times}3$, repeated twice, with skip connection.} \\
\multicolumn{2}{l}{\footnotesize GN = Group Normalization layer with 16 groups} \\
\bottomrule
\end{tabular}
\end{subtable}
\vspace{0.2cm}

\caption{\textbf{Architecture of $F_D$.} Consolidation network applied to the concatenated features from all $N{+}1$ maps.  Resolution is constant at $64{\times}64$.}
\label{tab:arch_fd_v2}
\vspace{-25pt}
\end{table}

\noindent \textbf{Animation Network $E$.}
$E$ takes the deformation latent $\mathbf{D} \in \mathbb{R}^{64 \times 64 \times 96}$ and the expression code $\mathbf{w} \in \mathbb{R}^{51}$. The expression code is first expanded via an MLP to 128 dimensions, then tiled spatially to $64{\times}64$ and concatenated with $\mathbf{D}$. A $1{\times}1$ convolution fuses the concatenated input, followed by 6 residual blocks and 2 upsampling stages that produce the residual delta map $E(\mathbf{w}, \mathbf{D}) \in \mathbb{R}^{H_{UV} \times W_{UV} \times d_G}$. We describe the full architecture in Tab.~\ref{tab:arch_e}.  While Eq.~\eqref{eq:residual} in the main paper describes the residual formulation conceptually in terms of activated Gaussian parameters, in practice we perform the addition in the \emph{pre-activation} parameter space before applying per-attribute activation functions. (see paragraph below for definition). We denote the raw (unactivated) canonical map as $\mathbf{C}_{\text{raw}}$, such that $\mathbf{C} = \mathrm{activate}(\mathbf{C}_{\text{raw}})$. To improve training convergence, we scale the delta by a small constant $\lambda_d = 0.01$:
\begin{equation}
  \hat{\mathbf{U}} = \mathbf{C}_\text{raw} +\Delta\mathbf{U}, \quad 
   \Delta\mathbf{U} = \lambda_d \cdot E(\mathbf{w,D})
  \label{eq:residual_scaling}
\end{equation}
where $\hat{\mathbf{U}}$ is the combined unactivated Gaussian UV map that is subsequently passed through per-attribute activations to produce the final Gaussian UV map $\mathbf{U} = \mathrm{activate}(\hat{\mathbf{U}})$ of Eq.~\eqref{eq:residual} in the main paper. This scaling ensures that in the early stages of training, $\hat{\mathbf{U}} \approx \mathbf{C}_{\text{raw}}$, effectively passing through the canonical frame's details (e.g. skin texture, freckles, hair) while the network gradually learns to predict expression-dependent offsets. We find that this significantly improves both detail preservation and convergence speed.

\begin{table}[h!]
\begin{tabular}{lcc}
\toprule
Type & Channels & Resolution \\
\midrule
Input $\mathbf{w}$            & 51              & --- \\
MLP (FC$\to$ReLU$\to$FC)     & $51 \to 128$    & --- \\
Tile to spatial               & 128             & $64^2$ \\
Concat ($\mathbf{D}$, tiled $\mathbf{w}$) & $96{+}128 \to 224$ & $64^2$ \\
Conv $1{\times}1$             & $224 \to 128$   & $64^2$ \\
ResBlock $\times 6$           & $128 \to 128$   & $64^2$ \\
UpBlock                       & $128 \to 64$    & $64^2 \to 128^2$ \\
UpBlock                       & $64 \to 32$     & $128^2 \to 256^2$ \\
Conv $3{\times}3$             & $32 \to 23$     & $256^2$ \\
\bottomrule
\multicolumn{3}{l}{\footnotesize ResBlock = Conv $3{\times}3$$\to$BN$\to$ReLU$\to$Conv $3{\times}3$$\to$BN, with skip connection.} \\
\multicolumn{3}{l}{\footnotesize UpBlock = Upsample$_2$$\to$Conv $3{\times}3$$\to$BN$\to$ReLU.} \\
\bottomrule
\end{tabular}
\vspace{5pt}
\caption{\textbf{Architecture of $E$} (animation network).}
\label{tab:arch_e}
\vspace{-25pt}
\end{table}

\newpage

\noindent \textbf{Gaussian Activations.}
\label{app:activations} After applying the residual predicted by $\mathbf{E}$ to $\mathbf{C}_\text{raw}$, we map the raw predictions to  Gaussian parameters using the same activation functions as in~\cite{headsup}. For each spatial location $(u,v)$ in the UV map:

\begin{align}
\boldsymbol{\mu}_{u,v} &=  \hat{\mathbf{U}}^{(\mu)}_{u,v},  \label{eq:act_pos} \\
\mathbf{s}_{u,v} &= \exp\!\left(\hat{\mathbf{U}}^{(s)}_{u,v}\right) \label{eq:act_scale} \\
\mathbf{q}_{u,v} &= \mathrm{normalize}\!\left(\hat{\mathbf{U}}^{(q)}_{u,v}\right) \label{eq:act_rot} \\
\alpha_{u,v} &= \sigma\!\left(\hat{\mathbf{U}}^{(\alpha)}_{u,v}\right) \label{eq:act_opacity} \\
\mathbf{k}_{u,v} &= \hat{\mathbf{U}}^{(k)}_{u,v}, \label{eq:act_color}
\end{align}
where $\sigma$ denotes the sigmoid function. Positions are anchored to a template mesh with an offset. Scales and opacities are mapped to positive values via exponential and sigmoid, respectively. Rotations are normalized to unit quaternions, and spherical harmonic coefficients are left unconstrained.

\noindent \textbf{Passthrough Network.}
The passthrough network takes a latent $\mathbf{Z}_i \in \mathbb{R}^{32 \times 32 \times 512}$ and produces the correction $\Delta\mathbf{P}_i \in \mathbb{R}^{256 \times 256 \times d_G}$ of Eq.~\eqref{eq:passthrough} in the main paper. It has three stages (Tab.~\ref{tab:arch_passthrough}): a residual convolutional encoder followed by an MLP that compresses the input to a 128-dimensional narrow bottleneck; a latent-fusion module that combines the bottleneck with the detached deformation latent $\mathbf{D}$ to produce 256-channel features at $64{\times}64$; and a residual convolutional decoder that upsamples to the Gaussian UV resolution.

\begin{table}[h!]
\centering
\begin{tabular}{lcc}
\toprule
Type & Channels & Resolution \\
\midrule
\multicolumn{3}{l}{\textit{Encoder + MLP bottleneck}} \\
Input ($\mathbf{Z}_i$)         & $512$               & $32^2$ \\
DownBlock                      & $512 \to 512$       & $32^2 \to 16^2$ \\
DownBlock                      & $512 \to 256$       & $16^2 \to 8^2$ \\
DownBlock                      & $256 \to 256$       & $8^2 \to 4^2$ \\
Flatten                        & $4096$              & --- \\
FC + LReLU                     & $4096 \to 256$      & --- \\
FC + LReLU                     & $256 \to 256$       & --- \\
FC                             & $256 \to 128$       & --- \\
\midrule
\multicolumn{3}{l}{\textit{Latent fusion (with deformation latent $\mathbf{D}$)}} \\
Tile bottleneck spatially      & $128$               & $64^2$ \\
Concat with $\mathbf{D}$       & $128{+}96 \to 224$  & $64^2$ \\
Conv $1{\times}1$              & $224 \to 256$       & $64^2$ \\
\midrule
\multicolumn{3}{l}{\textit{Residual decoder}} \\
Input                          & $256$               & $64^2$ \\
UpBlock                        & $256 \to 256$       & $64^2 \to 128^2$ \\
UpBlock                        & $256 \to 128$        & $128^2 \to 256^2$ \\
Conv $3{\times}3$              & $128 \to d_G$        & $256^2 \to 256^2$ \\
\bottomrule
\multicolumn{3}{l}{\footnotesize DownBlock = Conv $3{\times}3$ s2 $\to$ BN $\to$ ReLU.} \\
\multicolumn{3}{l}{\footnotesize UpBlock = Upsample$_2$$\to$Conv $3{\times}3$$\to$BN$\to$ReLU.} \\
\bottomrule
\end{tabular}
\vspace{5pt}
\caption{\textbf{Architecture of the passthrough network.}}
\label{tab:arch_passthrough}
\end{table}

\subsection{Training Setup}
\label{app:training_details}
For pretraining, we sample \num{100} expression-diverse frames per subject across all 10K identities. Evaluation is performed on \num{100} frames from a held-out set of \num{50} subjects. The frozen HeadsUp network, our newly trained modules, and the reimplemented Avat3r and FastGHA baselines all use the same train/val splits. The dataset is not publicly available. All participants gave written informed consent for the use of their data. Subjects shown in this paper provided additional explicit consent for publication of their images. Tab.~\ref{tab:training_details} summarizes the hyperparameters for both stages.

\begin{table}[h!]
\centering

\begin{tabular}{lcc}
\toprule
 & Pretraining & Finetuning \\
\midrule
Dataset                & Internal10K & Ava-256 \\
Image size ($H{\times}W$) & $500{\times}375$ & $512{\times}334$ \\
UV map size ($H_{UV}{\times}W_{UV}$) & $256{\times}256$ & $256{\times}256$ \\
HeadsUp ($H_E$, $H_D$)& frozen      & finetuned on Ava-256 \\
Trained modules        & $F_C$, $F_D$, $E$ & $F_C$, $F_D$, $E$ \\
Steps                  & 200K        & 15K \\
Input frames ($N$)     & 4           & 4 \\
Input sampling         & random      & furthest distance \\
Target frames          & 1 (random)  & 1 (random) \\
Batch size             & 32          & 32 \\
Optimizer              & Adam        & Adam \\
Learning rate          & $2{\times}10^{-4}$ & $2{\times}10^{-4}$ \\
Schedule               & constant    & constant \\
Precision              & fp32        & fp32 \\
Hardware               & $8{\times}$ A100 80GB & $8{\times}$ A100 80GB \\
Training time          & ${\sim}$3 days & ${\sim}$1 day \\
\midrule
\multicolumn{3}{l}{\textit{Loss weights (both stages)}} \\
$\lambda_1$            & \multicolumn{2}{c}{1.0} \\[0.1cm]
$\lambda_{\text{ssim}}$& \multicolumn{2}{c}{100} \\[0.1cm]
$\lambda_{\text{lpips}}$& \multicolumn{2}{c}{10} \\[0.1cm]
$\lambda_{\text{region}}$& \multicolumn{2}{c}{20} \\[0.1cm]
$\lambda_{\text{reg}}$ & \multicolumn{2}{c}{$10^{-2}$} \\[0.1cm]
 $\lambda^\text{pass}_{1}$       & \multicolumn{2}{c}{8} \\[0.1cm]
  $\lambda^\text{pass}_{\text{lpips}}$ & \multicolumn{2}{c}{80} \\
\bottomrule
\end{tabular}
\vspace{5pt}
\caption{\textbf{Training hyperparameters.}}
\label{tab:training_details}
\end{table}


\subsection{Training Losses}
\label{app:losses_details}
The loss $\mathcal{L}$ is defined as follows:
\begin{equation*}
\mathcal{L} = \lambda_1 \mathcal{L}_1 + \lambda_{\text{ssim}} \mathcal{L}_{\text{ssim}} + \lambda_{\text{lpips}}
  \mathcal{L}_{\text{lpips}} + \lambda_{\text{region}} \mathcal{L}_{\text{region}} + \lambda_{\text{reg}} \mathcal{L}_{\text{reg}} + \mathcal{L}_{\text{pass}}.
  \label{eq:total_loss_app}
\end{equation*}

\noindent \textbf{Image Reconstruction Loss.}
We use L1 and structural similarity loss between the rendered images:
\begin{equation}
\mathcal{L}_1 = \| \hat{\mathbf{I}} - \mathbf{I}_{GT} \|_1, \quad
\mathcal{L}_{\text{ssim}} = 1 - \text{SSIM}(\hat{\mathbf{I}}, \mathbf{I}_{GT}).
\label{eq:l1_loss}
\end{equation}

\noindent \textbf{Perceptual Loss.}
An LPIPS perceptual loss~\cite{zhang2018unreasonable} on the rendered image:
\begin{equation}
  \mathcal{L}_{\text{lpips}} = \sum_{l} \| \phi_l(\hat{\mathbf{I}}) - \phi_l(\mathbf{I}_{GT}) \|_1,
\end{equation}
where $\phi_l$ denotes the feature map at layer $l$ of a pretrained AlexNet\cite{krizhevsky2012imagenet} network.
  

\noindent \textbf{Gaussian Regularization Loss.}
To encourage Gaussians to stay opaque and have reasonable scales, we apply opacity and scales regularization losses similar to~\cite{saito2024rgca}:
\begin{equation}
  \mathcal{L}_{\text{reg}} = \mathrm{mean}(\ell_{reg}), \quad \ell_{\text{reg}} = \| 1 - \alpha \|_1 + 
  \begin{cases}
    1 / \max(\mathbf{s},\, 10^{-7}) & \text{if } \mathbf{s} < 0.1, \\
    (\mathbf{s} - 10.0)^2 & \text{if } \mathbf{s} > 10, \\
    0 & \text{otherwise}
  \end{cases}
  \label{eq:reg_loss}
\end{equation}
where $\alpha$ and $\mathbf{s}$ denote the opacity and scale of splats in $\mathbf{U}$.

\noindent \textbf{Region-based losses.}
To increase the gradient signal on expressive regions, we apply L1 and LPIPS reconstruction losses on cropped eye and mouth regions:
\begin{equation}
  \hat{\mathbf{I}}^{\text{eyes}},\, \hat{\mathbf{I}}^{\text{mouth}} = \text{crop}(\hat{\mathbf{I}},\, \mathbf{p}), \quad
  \mathbf{I}^{\text{eyes}}_{GT},\, \mathbf{I}^{\text{mouth}}_{GT} = \text{crop}(\mathbf{I}_{GT},\, \mathbf{p}),
  \label{eq:region_crops}
\end{equation}
where $\mathbf{p}$ denotes 2D facial landmarks detected in the target frame.



\subsection{Baselines}
\label{app:baselines_reimp}
For our baseline comparison, we evaluate our method against Avat3r~\cite{kirschstein2025avat3r} and FastGHA~\cite{ji2026fastgha}, representing state-of-the-art methods in feed-forward 3D head animation. For both baselines, we first pretrain on Internal10K and finetune on Ava-256, following the same two-stage protocol as our method. On Ava-256, we also compare against the concurrent UIKA~\cite{wu2026uika}.\\

\noindent \textbf{Avat3r~\cite{kirschstein2025avat3r}.}
Since the official Avat3r code is not available, we built upon the reimplementation used by the authors of HeadsUp~\cite{headsup}, provided to us. This reimplementation follows the corresponding publications as closely as possible. The original method precomputes DUSt3R~\cite{wang2024dust3r} and Sapiens~\cite{khirodkar2024sapiens} features only for a subset of \num{10} frames per subject due to the computational cost of running DUSt3R, as it operates on pairs of images, introducing $O(n^2)$ DUSt3R evaluations. Our setup requires all \num{1000} frames per subject, making DUSt3R extraction prohibitive. We replace it with DINOv3~\cite{simeoni2025dinov3}, a universal feature extractor which operates on single images and scales to per-frame extraction while retaining geometric awareness. For Sapiens~\cite{khirodkar2024sapiens}, we took a publicly available 1B checkpoint. On top of the HeadsUp implementation that does not require rigging with the same model, here we add expression conditioning, the way it is proposed in the original Avat3r work. Specifically, we transform expression codes through a shallow MLP into 20 tokens fed to the generator's cross-attention layers, following the Avat3r recipe, using the same expression codes as our method for driving. Confidence-based masking in skip connections of Avat3r has been implemented through known foreground masks that already give us foreground/background separation and remove the need to estimate it. On top of L1, SSIM, and LPIPS used in Avat3r, we also add DINOv2~\cite{oquab2023dinov2} perceptual loss similar to~\cite{oroz2025perchead} that, as we found, significantly improved the results. Additionally, we found it important to introduce two regularizations that significantly improve initialization and convergence: a scale penalty to prevent Gaussians from becoming too large and a silhouette loss that encourages the Gaussians opacity to be close to the target foreground mask (only enabled during the first few epochs). While DINOv2 loss and these additional regularizations were found to improve performance compared to literal reimplementation, we added them to reach better photorealism of Avat3r than reported, which makes that baseline even stronger. Without those changes, our Avat3r implementation is still producing plausible results. \\

\noindent \textbf{FastGHA~\cite{ji2026fastgha}.} FastGHA predicts pixel-aligned Gaussians with a feed-forward transformer conditioned on DINOv3~\cite{simeoni2025dinov3} and Stable Diffusion VAE features~\cite{rombach2022high}. To support sparse-view reenactment, Gaussians are predicted in the canonical head frame and rigged with per-Gaussian deformations predicted from expression codes and parameterized by a dedicated shallow MLP. FastGHA relies on position maps from VGGT \cite{wang2025vggt} as geometry supervision that acts as a regularization, unlike in Avat3r using similar feature matching maps from DUSt3R~\cite{wang2024dust3r} as an additional input. Precomputing VGGT maps on sets of multi-view images would require infeasible compute time, as it would require $O(n^d)$ evaluations ($n$ being the overall number of frames and $d$ the number of input views to VGGT), and it was omitted for that reason. However, this component should have a smaller effect on the final result, due to its regularization nature. Since the code of FastGHA is also not available publicly, similar to Avat3r, we relied on a FastGHA reimplementation that fully builds upon the one created by HeadsUp~\cite{headsup} authors. 
Following HeadsUp, we made two modifications to improve the convergence and generalization of FastGHA. First, we added an extra learnable encoder in parallel to the frozen DINOv3~\cite{simeoni2025dinov3} and Stable Diffusion VAE~\cite{rombach2022high} encoders. Second, we replaced the frozen Stable Diffusion VAE decoder with a learnable residual decoder to improve gradient.

Training is stage-wise: in the beginning of training, we train FastGHA for a number of epochs in a reconstruction-only setting (i.e. when for every individual sample, inputs and targets are always from the same timestamp -- as in HeadsUp), and only then continue in standard reenactment setting, as we observed the optimization was not as stable if training directly for reenactment. Also, we introduce a silhouette loss (during the first few epochs) and a scale penalty for better convergence and initialization, the same way as for Avat3r. Incorporated changes only marginally improve quality and were added to make FastGHA an even stronger baseline on top of the literal reimplementation. \\

\noindent \textbf{UIKA~\cite{wu2026uika}.} UIKA reconstructs an animatable Gaussian head avatar from a given number of input images in a single forward pass. It reprojects each input image into the FLAME UV space, fuses all views with learnable UV tokens in a transformer, and decodes them into canonical Gaussians that are animated with linear blend skinning. Like our method, it builds a fixed-size UV representation that is independent of the number of input views and animates in real time. This makes it the most closely related concurrent baseline. We start from the authors' released checkpoint that was trained on public monocular and multi-view head datasets. This checkpoint takes the role of the Internal10K pretraining used for the other baselines and keeps UIKA's prior public and reproducible. We finetune it on Ava-256 with the official training code and use the same training data and held-out subjects as for our method. We keep UIKA's architecture, frozen DINOv3 encoder, losses and optimizer unchanged. Unlike our method, UIKA is driven by FLAME parameters instead of ARKit codes. We therefore add a small MLP, trained jointly with UIKA, to map our ARKit codes to FLAME~\cite{li2017flame} expression, jaw, eye and neck parameters. The head pose is given by the cameras, expressed in the per-frame head coordinate frame. To bring the Ava-256 data into UIKA's canonical FLAME space, we register the FLAME template to the Ava-256 mesh template and fit per-subject FLAME shape parameters to the tracked Ava-256 meshes. Each training sample uses 1 to 4 input views and 4 supervision views. We train on 2 H100 GPUs with a batch size of 2 per GPU. We sweep the learning rate over $\{5\times10^{-5}, 10^{-4}, 5\times10^{-4}\}$ and keep $5\times10^{-5}$, as it gives the best LPIPS, PSNR and CSIM. We report the checkpoint at 8k iterations, after the metrics have plateaued. At test time, UIKA uses the same 4 input views as our method and renders its 134K Gaussians at the same $512\times334$ resolution as our method.


\newlength{\imgheight}
\setlength{\imgheight}{5cm}
  
\begin{figure}[h!]
    \centering
    \setlength{\tabcolsep}{2pt}
    \renewcommand{\arraystretch}{0}
    \begin{tabular}{c ccc}
        \rotatebox{90}{\makebox[\imgheight][c]{\footnotesize Avat3r from HeadsUp}} &
        \includegraphics[width=0.3\linewidth]{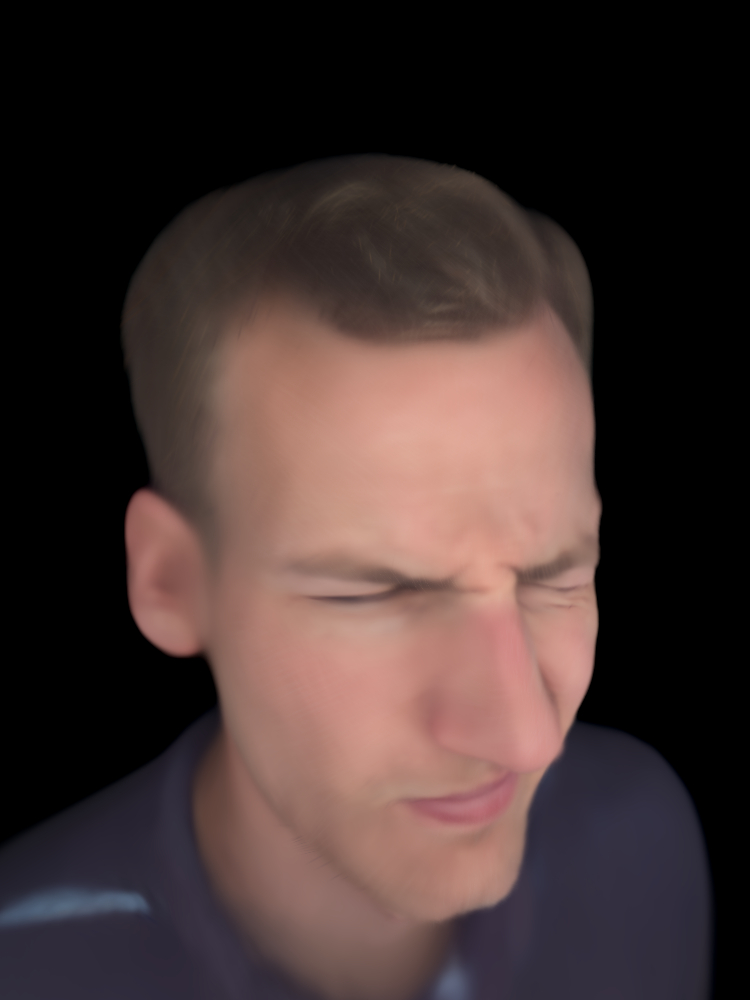} &
        \includegraphics[width=0.3\linewidth]{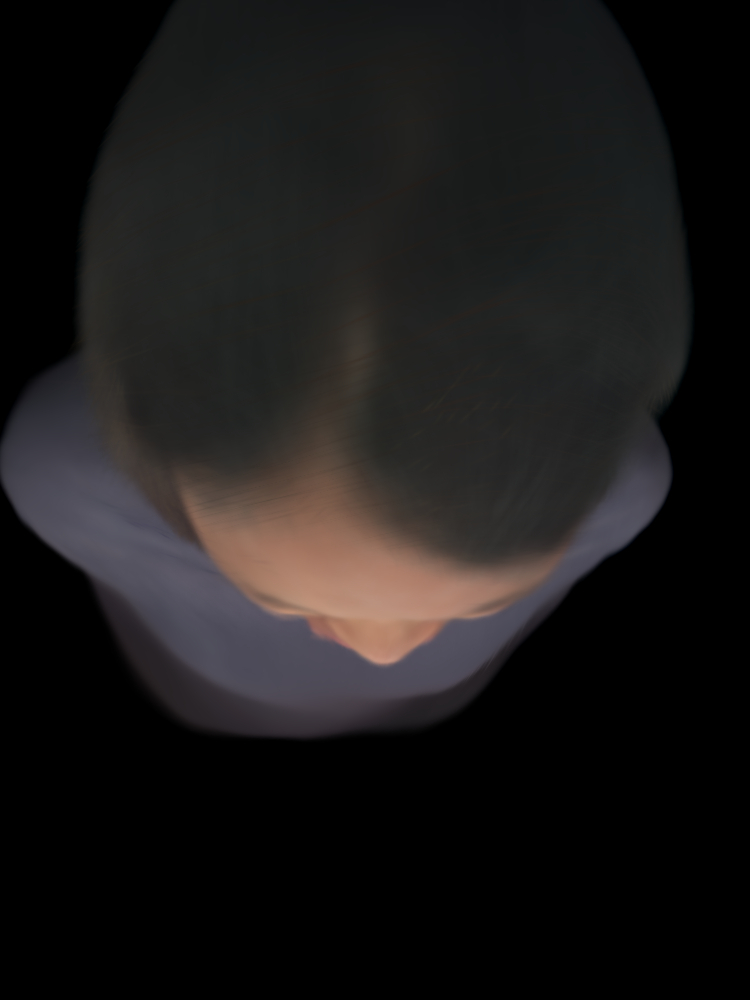} &
        \includegraphics[width=0.3\linewidth]{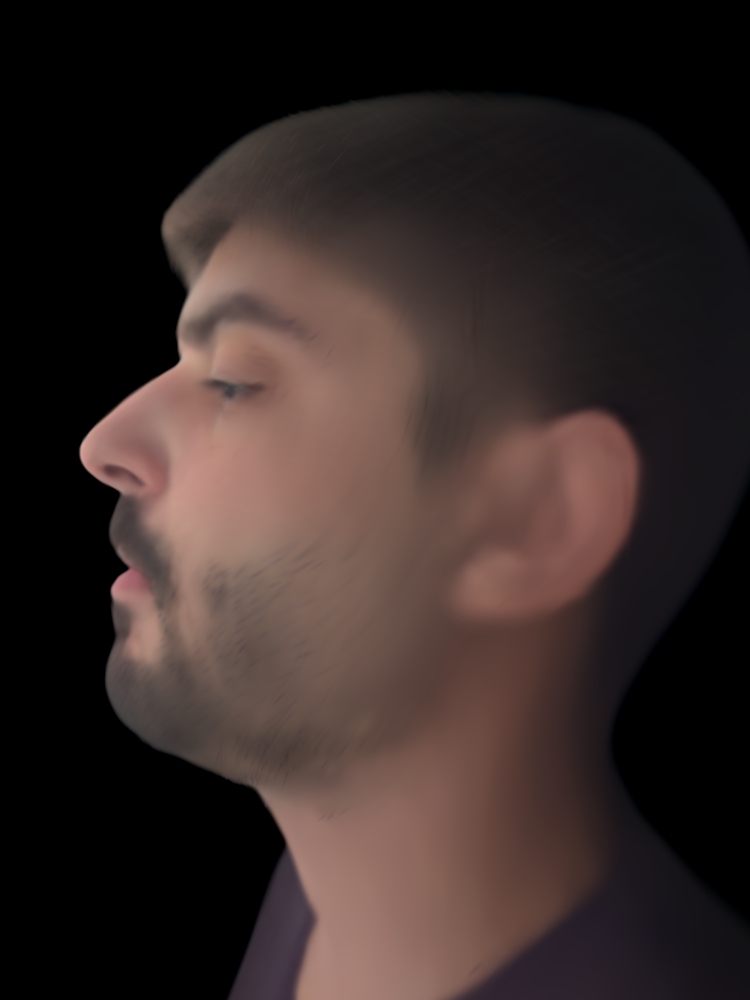} \\
        \rotatebox{90}{\makebox[\imgheight][c]{\footnotesize Avat3r Ours}} &
        \includegraphics[width=0.3\linewidth]{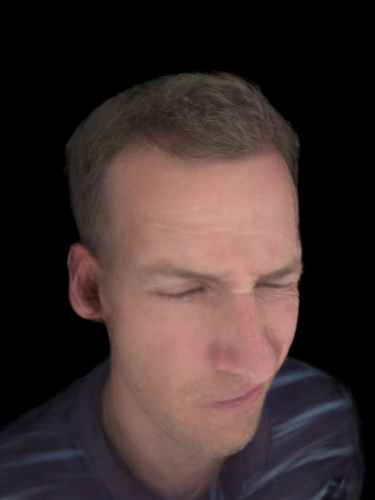} &
        \includegraphics[width=0.3\linewidth]{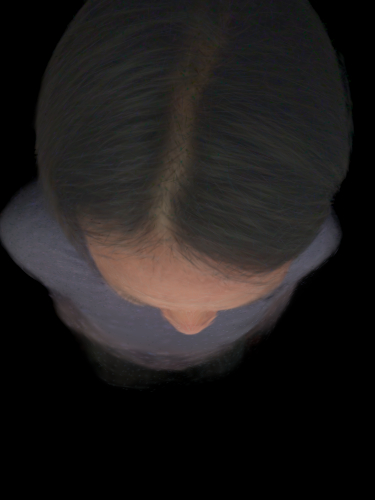} &
        \includegraphics[width=0.3\linewidth]{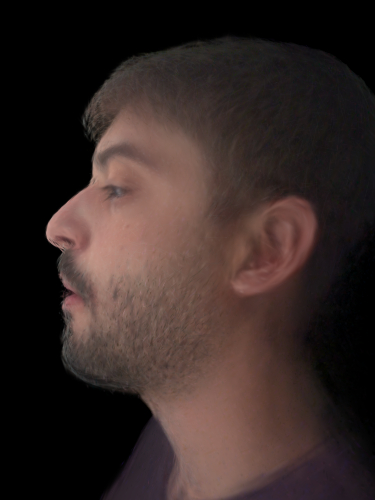} \\
    \end{tabular}
    \caption{Qualitative comparison between Avat3r implementation in HeadsUp and in our method in a setting without rigging.
    Main differences in our Avat3r implementation include DINOv2 loss (which explains sharper detail), as well as relaxing feature extractor choice from DUSt3R to DINOv3 (while keeping Sapiens in place), and training with expression conditioning.}
    \label{fig:baselines_vs_headsup}
\end{figure}

\subsection{Body Alignment Network (BAN)}
\label{app:ban_details}
Fig.~\ref{fig:gt_body_align_architecture} in the main paper shows an overview of BAN training. This section provides more details on the network architecture and supervision losses.

\noindent \textbf{Blending network.} The blending network is a 6-layer convolutional network operating at $32{\times}32$ resolution. It takes the concatenated latents $[\mathbf{Z}_\text{ref}, \mathbf{Z}_\text{GT}] \in \mathbb{R}^{ 32 \times 32 \times 1024}$ and outputs a 1024-channel tensor split channel-wise into the blending weights $\mathbf{m}$ and correction $\boldsymbol{\delta}_a$. The final convolution is zero-initialized so the network starts at a $50/50$ average of $\mathbf{Z}_\text{ref}$ and $\mathbf{Z}_\text{GT}$.

\noindent \textbf{Training supervision.}
The aligned Gaussian UV map $\mathbf{A}$ is rendered from multiple viewpoints. Pretrained face-parsing and person-segmentation networks operate on the renderings to produce binary masks for the face region and the body region. The face region of the aligned rendering is supervised against the target rendering, while the body region is supervised against the reference rendering. Mask boundaries are randomized stochastically per iteration to prevent seam artifacts.

\noindent \textbf{Losses.}
L1 and LPIPS reconstruction losses are applied to the masked face and body regions, while an adversarial loss is applied to the full rendering (Tab.~\ref{tab:ban_loss_weights}). The adversarial loss is a hinge GAN applied on $256{\times}256$ random crops of the rendering. The discriminator is conditioned on a separate reference-view rendering to encourage subject-specific realism. A gradient penalty is applied periodically during training for stability.

\begin{table}[h!]
\centering
\begin{tabular}{lc}
\toprule
Loss & Weight \\
\midrule
$\mathcal{L}_1$              & 1.0 \\
$\mathcal{L}_{\text{lpips}}$ & 10.0 \\
$\mathcal{L}_{\text{adv}}$   & 0.25 \\
\bottomrule
\end{tabular}
\vspace{5pt}
\caption{\textbf{BAN loss weights.}}
\label{tab:ban_loss_weights}
\end{table}

\noindent \textbf{Training.}
BAN is trained for 20K iterations with Adam (learning rate $5\times 10^{-5}$, batch size 16).

\subsection{Body Temporal Stability Analysis}
\label{app:temporal_body_metrics}
Beyond the RGB body-stability metric $\sigma^t_\text{body}$ reported in the main paper, we report two variants on the same body-view renders: $\Delta E_{76}$ applies the identical per-pixel temporal-deviation computation in CIELAB rather than RGB, giving a perceptual color-flicker measure; and TCE (Temporal Consistency Error) measures geometric drift as the mean dense optical-flow magnitude (OpenCV DIS, in pixels) from each subject's first render to its others. Tab.~\ref{tab:temporal_supp} reports all three; they agree that Ours$_\text{+BAN}$ is the most stable.

\begin{table}[h!]
  \centering
  \setlength{\tabcolsep}{8pt}
  \begin{tabular}{lccc}
  \toprule
   & $\sigma^t_\text{body}$$\downarrow$ & $\Delta E_{76}$$\downarrow$ & TCE$\downarrow$ \\
  \midrule
  Avat3r             & 18.8            & 10.0 & 11.7 \\
  FastGHA            & 22.9            & 12.1 & 13.4 \\
  Ours               & ~\underline{6.0} & ~\underline{2.8} & ~\underline{2.8} \\
  Ours$_\text{+BAN}$ & ~\textbf{0.7}    & ~\textbf{0.2} & ~\textbf{0.2} \\
  \bottomrule
  \end{tabular}
  \vspace{5pt}
   \caption{\textbf{Body Temporal Stability on Ava-256.} Per-pixel temporal standard deviation of the body-view render in RGB~($\sigma^t_\text{body}$) and CIELAB~($\Delta E_{76}$), and optical-flow drift~(TCE, in pixels), computed across each subject's expressions; lower is more stable. Ours$_\text{+BAN}$ is the most stable across all three metrics. Best in \textbf{bold}, second best \underline{underlined}.}
  \label{tab:temporal_supp}
  \end{table}

\subsection{Stable \vs Frozen Body}
\label{app:frozen_body}
\begin{figure}[h!]
  \centering
  \makebox[\linewidth]{%
    \makebox[0.25\linewidth]{\footnotesize Body Source}%
    \makebox[0.25\linewidth]{\footnotesize Ours$_\text{+BAN}$}%
    \makebox[0.25\linewidth]{\footnotesize Frozen Body}%
    \makebox[0.25\linewidth]{\footnotesize Ours}}\\[1pt]
  \includegraphics[width=\linewidth, trim=0 200 0 0, clip]{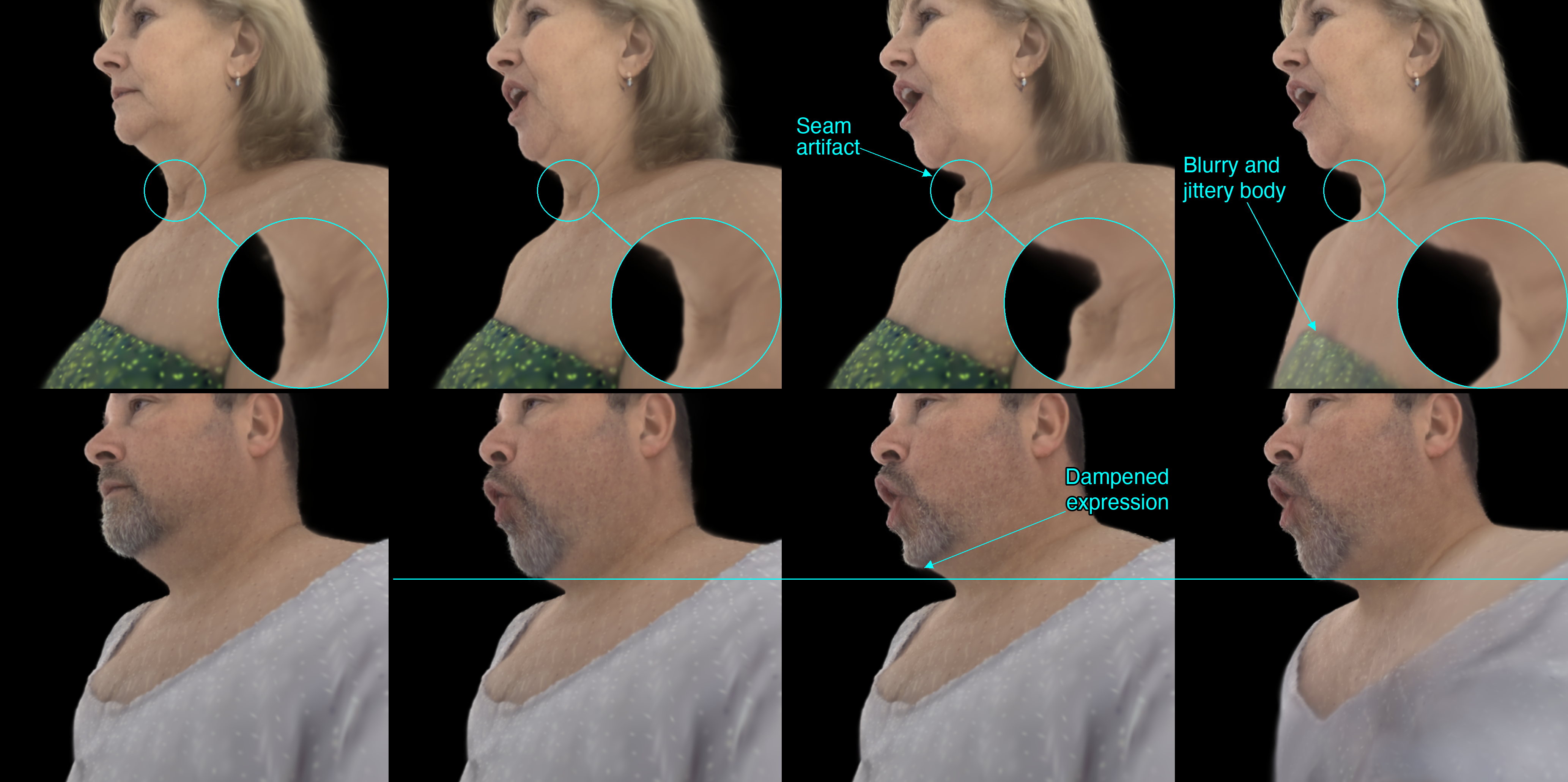}
  \caption{\textbf{Frozen-body blending \vs BAN.} At a mouth opening, a narrow ramp leaves a seam at the neck (top), and a wider ramp changes the chin shape and reduces expression fidelity (bottom). Ours$_\text{+BAN}$ avoids both.}
  \label{fig:frozen_body}
\end{figure}

A low $\sigma^t_\text{body}$ alone cannot separate a stable body from a frozen one: a frozen body gives $\sigma^t_\text{body}=0$, whereas Ours$_\text{+BAN}$ gives $0.7$ (Tab.~\ref{tab:Viewsexperiments}) and retains realistic neck motion (see Supp. videos). A natural alternative to BAN is to freeze the body at inference time. We tried this by blending a frozen body Gaussian UV map from a single frame into the animated Gaussian UV map of Ours with a smooth ramp across the neck. As shown in Fig.~\ref{fig:frozen_body}, a narrow ramp leaves a seam at the neck when the mouth opens, while widening the frozen and blended range to hide the seam alters the chin shape and reduces expression fidelity. BAN avoids this trade-off: since the body is aligned in the supervision targets (Sec.~\ref{sec:body-alignment}), the animation network predicts a single consistent Gaussian UV map without any blending boundary.

\subsection{Runtime Performance} \label{sec:suppRuntime}

Tab.~\ref{tab:runtime_app} shows additional animation runtime analysis on H100 and iPhone 17 Pro (GPU) for 1, 2, and 4 input views. As demonstrated in Tab.~\ref{tab:merged_quality_runtime} in the main paper and Tab.~\ref{tab:runtime_app}, the animation time for our method remains constant with the number of input views unlike Avat3r and FastGHA. For \num{1} view, our method is \num{2}$\times$ faster than FastGHA on H100 and \num{3.5}$\times$ faster on iPhone 17 Pro GPU while it is \num{7.7}$\times$ faster for \num{4} views than FastGHA on H100 and \num{13}$\times$ faster on iPhone 17 Pro GPU.  

\begin{table}[h!]
\centering
\begin{tabular}{llcccccc}
\toprule
   & & \multicolumn{2}{c}{Avat3r~\cite{kirschstein2025avat3r}}
     & \multicolumn{2}{c}{FastGHA~\cite{ji2026fastgha}}
     & \multicolumn{2}{c}{Ours} \\
\cmidrule(lr){3-4} \cmidrule(lr){5-6} \cmidrule(lr){7-8}
Device & $N$ & ms & FPS & ms & FPS & ms & FPS \\
\midrule
\multirow{3}{*}{H100} & 1 & 21.33 & 47 & 2.39 & 418 & \textbf{1.18} & \textbf{851} \\
                      & 2 & 41.15 & 24 & 4.67 & 214 & \textbf{1.18} & \textbf{851} \\
                      & 4 & 78.44 & 13 & 9.14 & 109 & \textbf{1.18} & \textbf{851} \\
\midrule
\multirow{3}{*}{\shortstack[l]{iPhone 17 Pro\\(GPU)}}
                      & 1 & -- & -- & 17.6  & 57 & \textbf{5.1} & \textbf{196} \\
                      & 2 & -- & -- & 31.76 & 31 & \textbf{5.1} & \textbf{196} \\
                      & 4 & -- & -- & 66.88 & 15 & \textbf{5.1} & \textbf{196} \\
\bottomrule
\end{tabular}
\vspace{5pt}
\caption{\textbf{Animation runtime (ms) and FPS. }Additional results for animation runtime (ms) and FPS for our method, FastGHA, and Avat3r on different devices across varying numbers of input views ($N$).}
\label{tab:runtime_app}
\end{table}

\noindent\textbf{Number of Gaussians.} To disentangle the number of Gaussians from the animation architecture, we increase the number of Gaussians of our method by $4\times$, from 65K to 260K (Ours260). As shown in Tab.~\ref{tab:runtime_260}, Ours260 remains 2--3.5$\times$ faster than FastGHA at its fastest setting of 150K Gaussians.

\begin{table}[h!]
\centering
\begin{tabular}{lrcc}
\toprule
Method & \#G & A100 & iPhone 17 Pro (ANE) \\
\midrule
FastGHA~\cite{ji2026fastgha} ($N{=}1$) & 150K & 6.81 & 11.80 \\
Ours260 & 260K & 3.3 & 3.4 \\
Ours    & \textbf{65K} & \textbf{1.92} & \textbf{1.45} \\
\bottomrule
\end{tabular}
\vspace{5pt}
\caption{\textbf{Animation runtime (ms) with $4\times$ more Gaussians (Ours260).}}
\label{tab:runtime_260}
\end{table}

\noindent\textbf{End-to-End Latency.} Tab.~\ref{tab:runtime_e2e} reports the complete online latency. Rendering with gsplat~\cite{ye2025gsplat} on an A100 takes 1.91\,ms for Avat3r, 1.53\,ms for FastGHA and 1.33\,ms for our method, giving end-to-end latencies of 168.14\,ms, 27.78\,ms and 3.25\,ms. The end-to-end latency is dominated by the animation stage.
As a reference point, the end-to-end latency reported in LAM~\cite{he2025lam} (which is reported to be significantly behind in quality compared to the concurrent UIKA~\cite{wu2026uika}) is 3.56\,ms.

\begin{table}[h!]
\centering
\begin{tabular}{lccc}
\toprule
 & Animation & Rendering & End-to-end \\
\midrule
Avat3r~\cite{kirschstein2025avat3r} & 166.23 & 1.91 & 168.14 \\
FastGHA~\cite{ji2026fastgha}        & 26.25  & 1.53 & 27.78  \\
\textit{LAM}$^\dagger$~\cite{he2025lam}                & - & - & \textit{3.56} \\
Ours                                & \textbf{1.92} & \textbf{1.33} & \textbf{3.25} \\
\bottomrule
\end{tabular}
\vspace{5pt}
\caption{\textbf{End-to-end online latency (ms)} on an A100 with $N{=}4$ input views. Rendering uses gsplat~\cite{ye2025gsplat}; animation times are from Tab.~\ref{tab:merged_quality_runtime}.
$^\dagger$Reported by the authors for their 80K-Gaussian model.
}
\label{tab:runtime_e2e}
\end{table}

\subsection{Further Number of Input Views Ablation} 
\label{sec:suppNViews}

\begin{figure*}[ht!]
  \centering
  \setlength{\tabcolsep}{0pt}
  \renewcommand{\arraystretch}{0}
  \footnotesize
  \begin{tabular}{@{}ccc@{\hspace{1mm}}ccc@{\hspace{1mm}}ccc@{\hspace{1mm}}c@{}}
  \multicolumn{3}{c}{$N=1$}  & \multicolumn{3}{c}{$N=2$}  & \multicolumn{3}{c}{$N=4$} & GT \\[4pt]
    Avat3r  & FastGHA  & Ours & Avat3r   & FastGHA  & Ours & Avat3r & FastGHA & Ours & \\[4pt]
\includegraphics[height=0.12\textwidth]{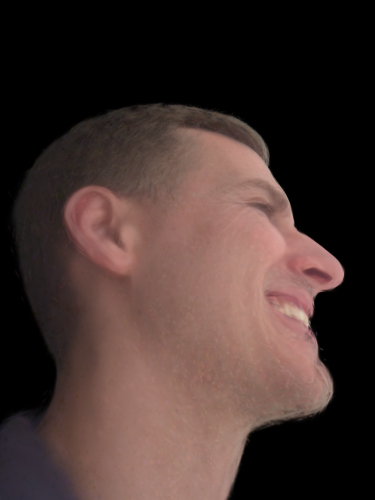} &
  \includegraphics[height=0.12\textwidth]{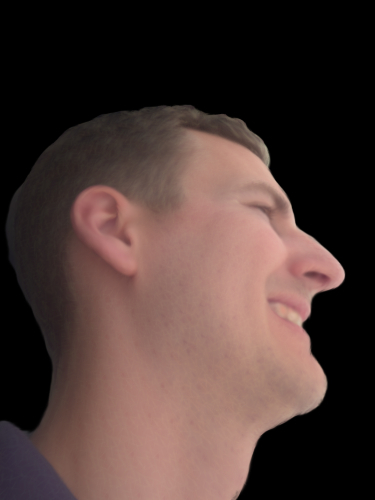} & 
  \includegraphics[height=0.12\textwidth]{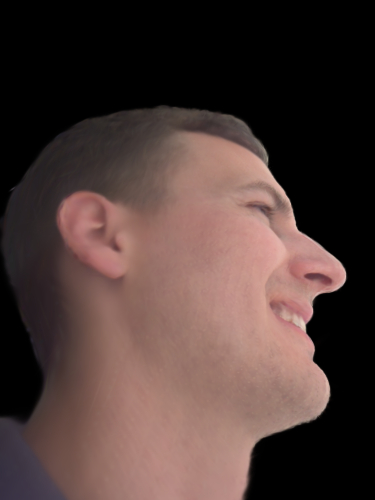} & 
  \includegraphics[height=0.12\textwidth]{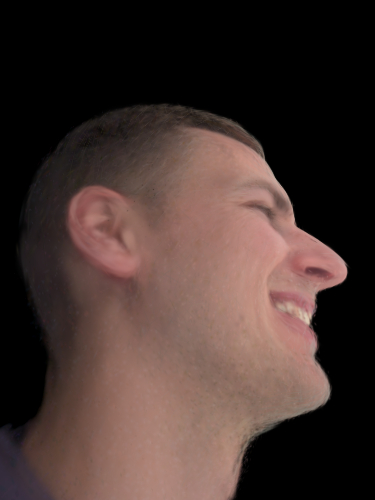} &
    \includegraphics[height=0.12\textwidth]{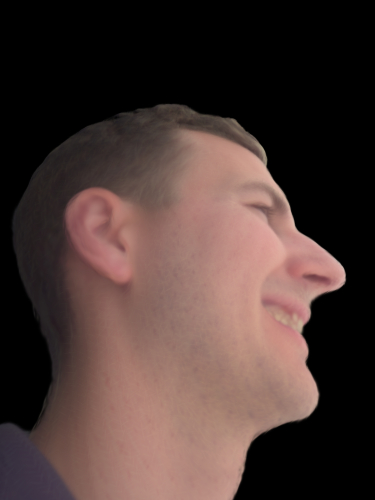} &
    \includegraphics[height=0.12\textwidth]{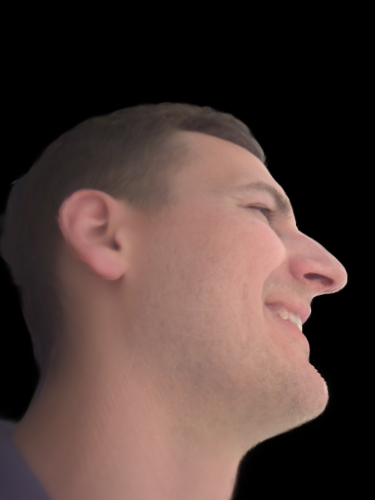} &
      \includegraphics[height=0.12\textwidth]{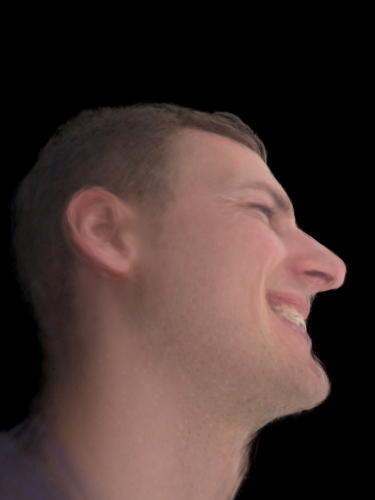} &
    \includegraphics[height=0.12\textwidth]{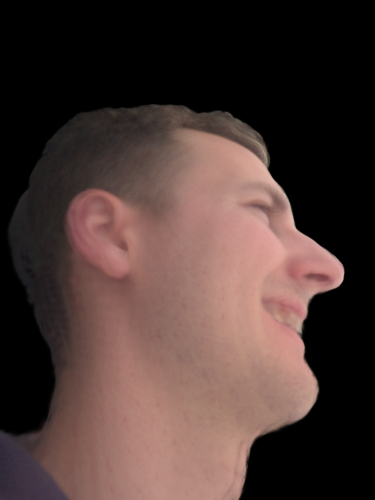} &
    \includegraphics[height=0.12\textwidth]{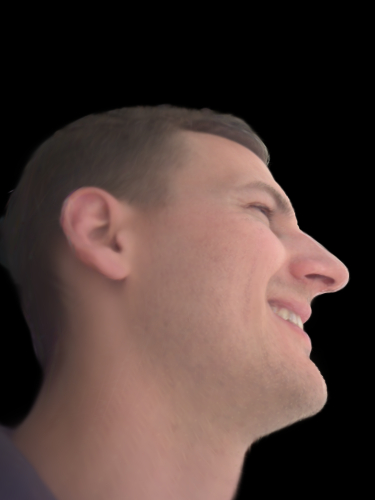} &
        \includegraphics[height=0.12\textwidth]{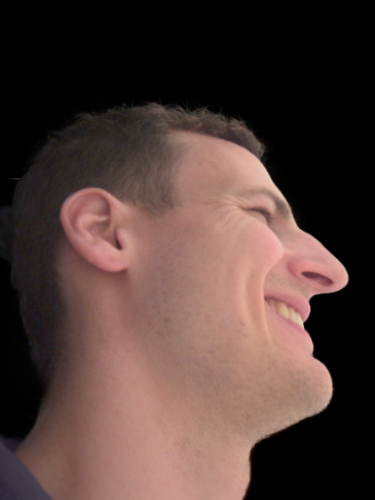} \\
\includegraphics[height=0.12\textwidth]{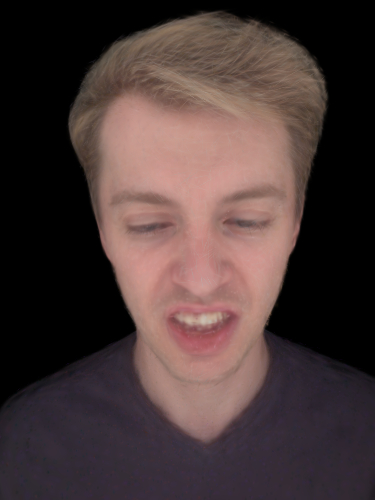} &
  \includegraphics[height=0.12\textwidth]{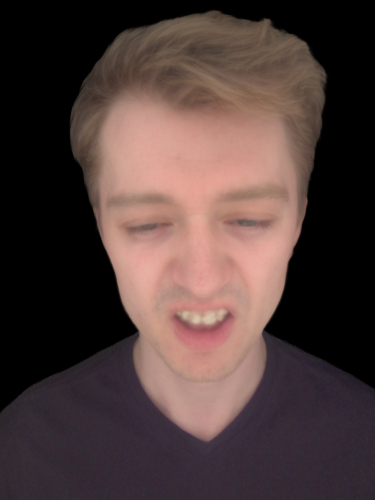} &
  \includegraphics[height=0.12\textwidth]{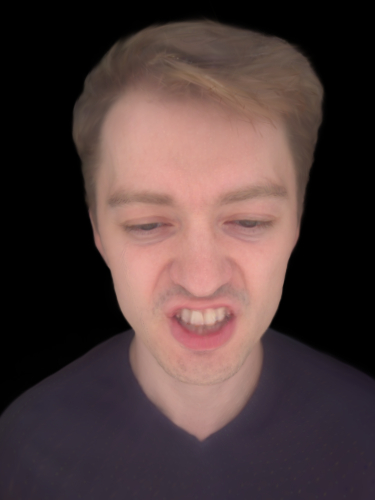} &
  \includegraphics[height=0.12\textwidth]{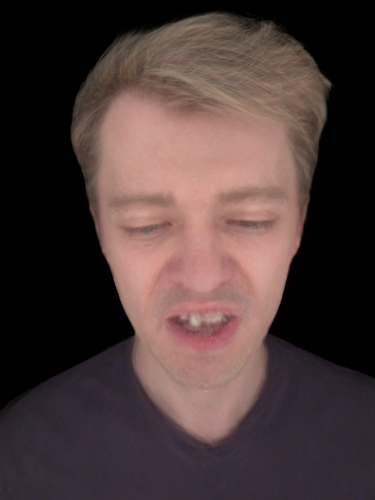} &
    \includegraphics[height=0.12\textwidth]{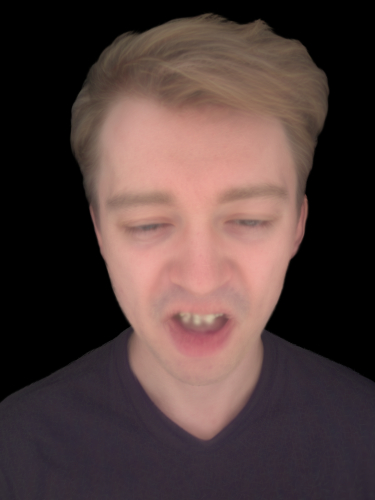} &
    \includegraphics[height=0.12\textwidth]{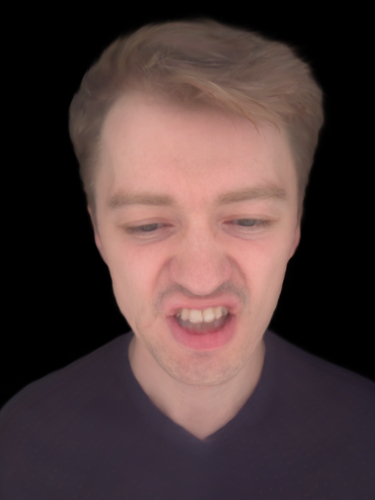} &
      \includegraphics[height=0.12\textwidth]{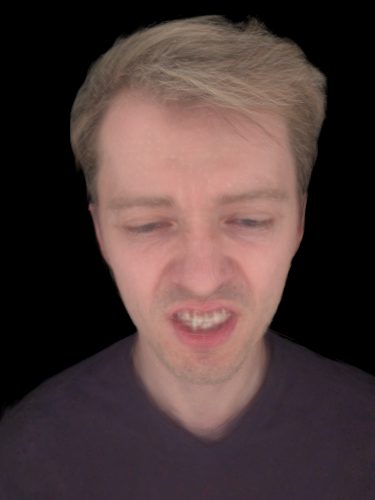} &
    \includegraphics[height=0.12\textwidth]{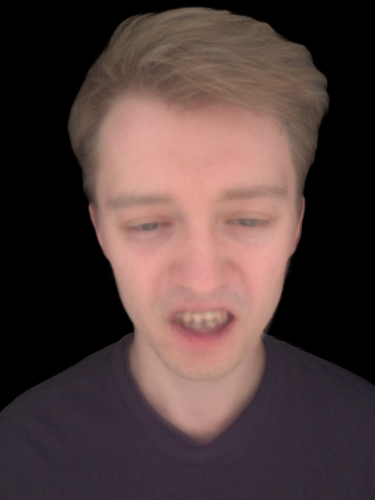} &
    \includegraphics[height=0.12\textwidth]{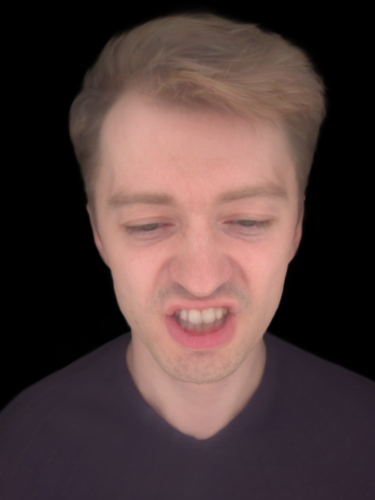} &
        \includegraphics[height=0.12\textwidth]{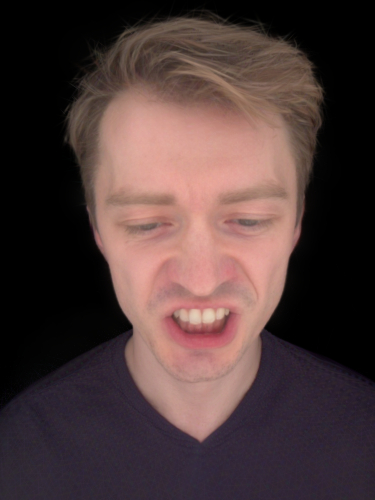} \\
  \end{tabular}
  \caption{\textbf{Impact of the number of input views.} Reconstruction quality scales naturally with the number of input images. Even with $N=1$, Ours produces sharper reconstruction and preserves identity better compared to prior work.}
  \label{fig:view_scaling}
  \end{figure*}

Fig.~\ref{fig:view_scaling} shows a qualitative comparison across numbers of input views.

\subsection{Prior Contribution}

To isolate the contribution of the prior from the animation network, we evaluated how a weaker prior affects quality in Tab.~\ref{tab:rebuttal_encoder_views}. Specifically, we trained our method using a HeadsUp prior trained on only 1k subjects (Ours1k) instead of 10k (Ours). While reconstruction quality decreases, it remains comparable on SSIM and CSIM with the Avat3r and FastGHA baselines in Tab.~\ref{tab:merged_quality_runtime}. Crucially, Ours1k still achieves an AKD of 7.8, outperforming Avat3r (8.3) and FastGHA (9.3), demonstrating that it retains high animation fidelity even under a constrained prior.

\begin{table}[h]
\centering
\begin{tabular}{lccccc}
\toprule
Method & LPIPS$\downarrow$ & SSIM$\uparrow$ & AKD$\downarrow$ & CSIM$\uparrow$ & PSNR$\uparrow$ \\
\midrule
Ours1k & 0.28 & 0.76 & 7.8 & 0.60 & 20.7 \\
Ours     & \textbf{0.24} & \textbf{0.77} & \textbf{7.4} & \textbf{0.67} & \textbf{21.6} \\
\bottomrule
\end{tabular}
\vspace{3pt}
\caption{\textbf{Ablation on weaker priors.} Impact of reducing the training data for the HeadsUp prior from 10k (Ours) to 1k (Ours1k) subjects.}
\label{tab:rebuttal_encoder_views}
\end{table}

\subsection{Identity Network Performance} \label{sec:suppIdentityNetworkPerformance}

We additionally compared the latency of our identity network to the analogous appearance-encoding stages of Avat3r~\cite{kirschstein2025avat3r} and FastGHA~\cite{ji2026fastgha}. In each method we measured all steps from the input images up to the decoding stage: for ours, up to the canonical Gaussian UV maps and deformation latents; for Avat3r, up to (but excluding) the cross-view transformer; for FastGHA, up to (but excluding) the Gaussian decoder. All measurements were performed on a single A100, using one batch of four input views.

\noindent \textbf{Measurements.}
Our identity network runs in 71\,ms, compared to 242\,ms for FastGHA and 2384\,ms for Avat3r. The identity network is executed only once per identity, while subsequent animation is significantly faster (see Tab.~\ref{tab:runtime_app}). For the baselines, the main bottlenecks are the image encoders: Sapiens~\cite{khirodkar2024sapiens} in Avat3r and DINOv3~\cite{simeoni2025dinov3}  in FastGHA (or even heavier VGGT). When we reduced the Sapiens input resolution in Avat3r from 1024$\times$1024 to 512$\times$512 or 512$\times$320, its latency dropped to 588\,ms and 437\,ms, respectively, which is still in both cases significantly larger than for our method.

\subsection{In-The-Wild Setting} \label{sec:suppITW}

\begin{figure*}[t]
  \centering
  \setlength{\tabcolsep}{1pt}
  \renewcommand{\arraystretch}{0.2}
  \newcommand{\ITWbase}{images/itw_self}
  \newcommand{\ITWimg}{0.125\linewidth}
  \newcommand{\ITWcropimg}[1]{%
    \adjustbox{trim=0 0.25\height{} 0 0.25\height, clip}{#1}%
  }
  \newcommand{\ITWfullimg}[1]{\includegraphics[width=\ITWimg]{#1}}
  \newcommand{\ITWlblshift}{20pt}
  \newcommand{\ITWlbllshift}{13pt}
  \newcommand{\ITWrowlbl}[1]{%
    \raisebox{\ITWlblshift}{\rotatebox[origin=c]{90}{\small #1}}%
  }
  \newcommand{\ITWrowlbll}[1]{%
    \raisebox{\ITWlbllshift}{\rotatebox[origin=c]{90}{\small #1}}%
  }
  \begin{tabular}{@{}c c c c c c c c@{}}
    \makebox[\ITWimg]{ Input}
    &
    & \multicolumn{6}{c}{ Self-reenactment} \\[2pt]

    \includegraphics[width=\ITWimg]{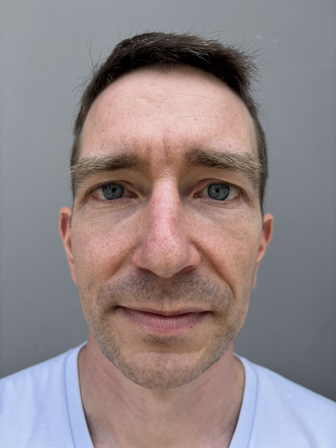}
    & \ITWrowlbl{Ours}
    & \includegraphics[width=\ITWimg]{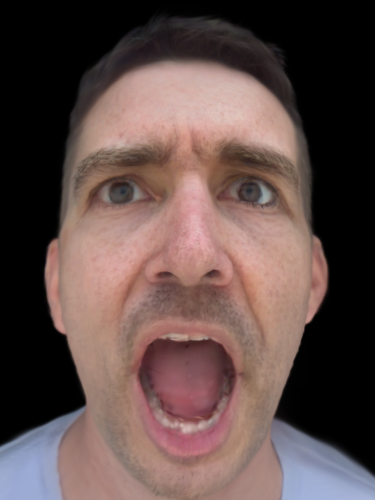}
    & \includegraphics[width=\ITWimg]{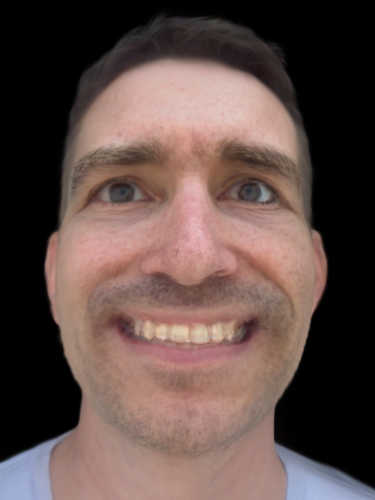}
    & \includegraphics[width=\ITWimg]{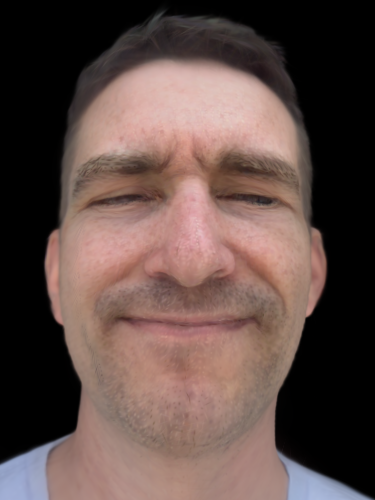}
    & \includegraphics[width=\ITWimg]{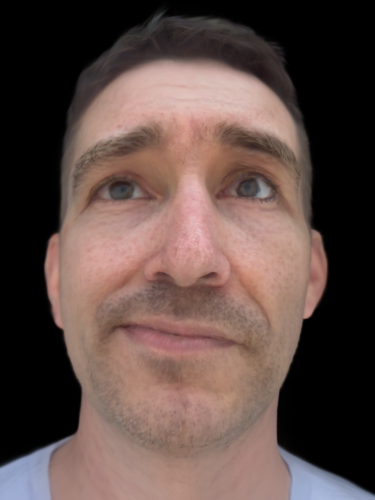}
    & \includegraphics[width=\ITWimg]{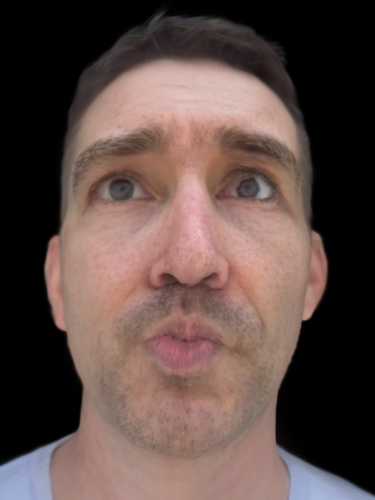}
    & \includegraphics[width=\ITWimg]{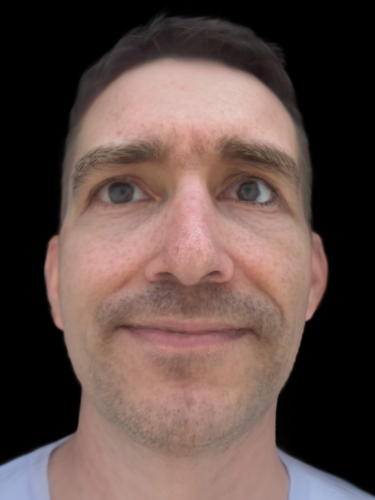} \\

    \includegraphics[width=\ITWimg]{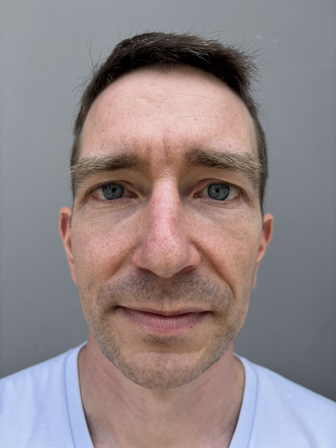}
    & \ITWrowlbl{Avat3r}
    & \includegraphics[width=\ITWimg]{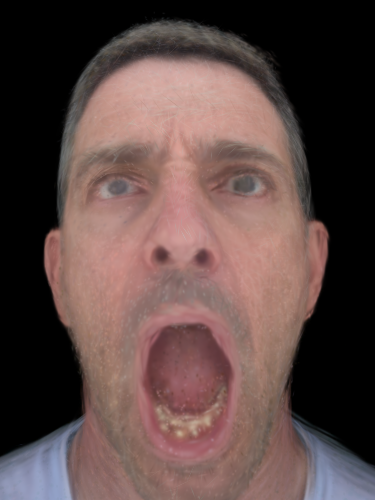}
    & \includegraphics[width=\ITWimg]{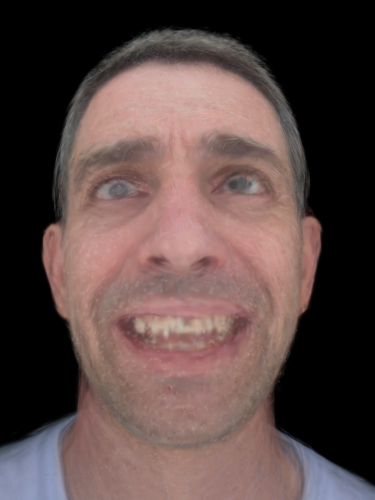}
    & \includegraphics[width=\ITWimg]{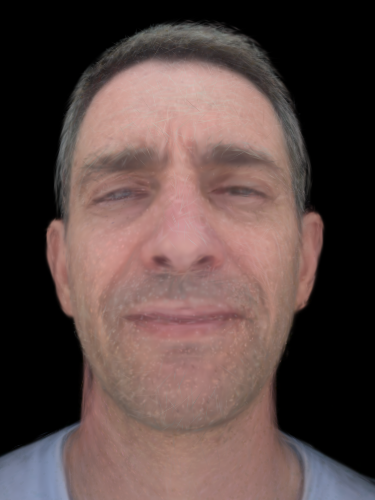}
    & \includegraphics[width=\ITWimg]{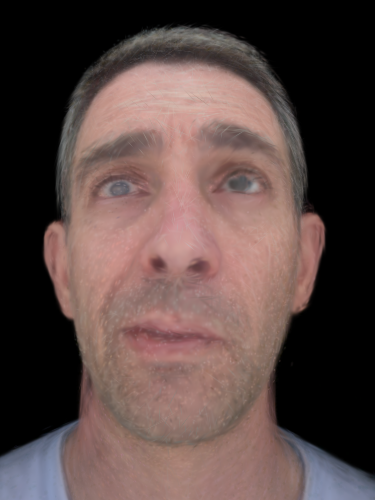}
    & \includegraphics[width=\ITWimg]{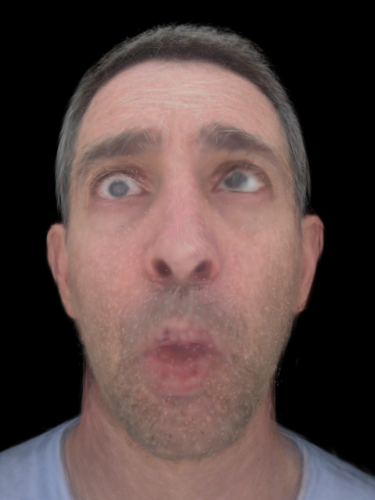}
    & \includegraphics[width=\ITWimg]{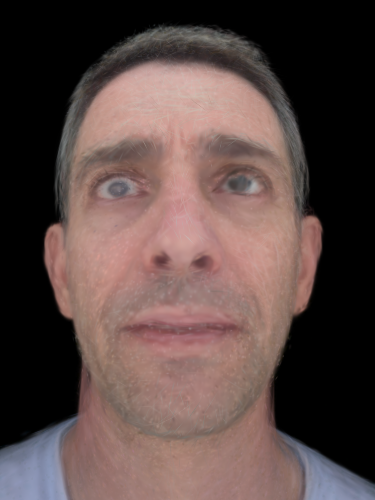} \\

    \includegraphics[width=\ITWimg]{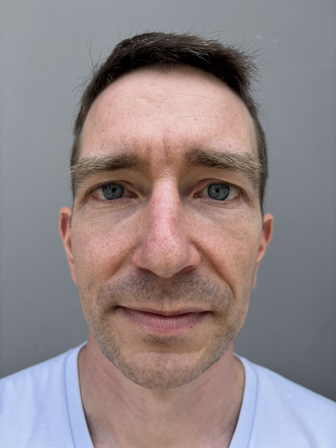}
    & \ITWrowlbl{FastGHA}
    & \includegraphics[width=\ITWimg]{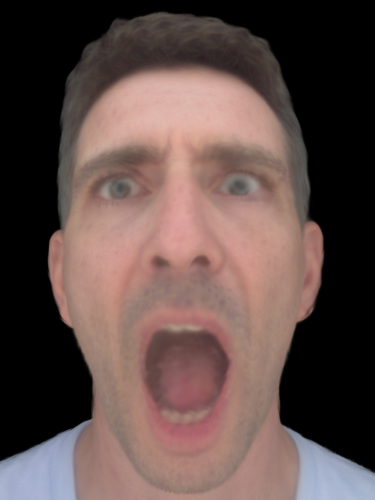}
    & \includegraphics[width=\ITWimg]{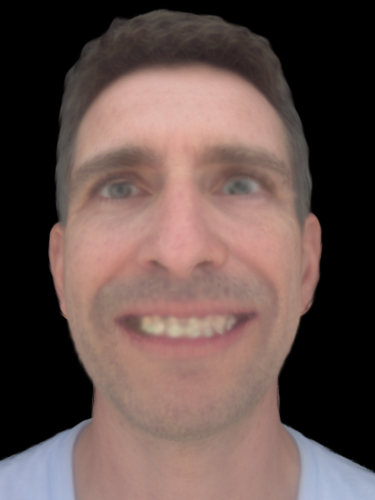}
    & \includegraphics[width=\ITWimg]{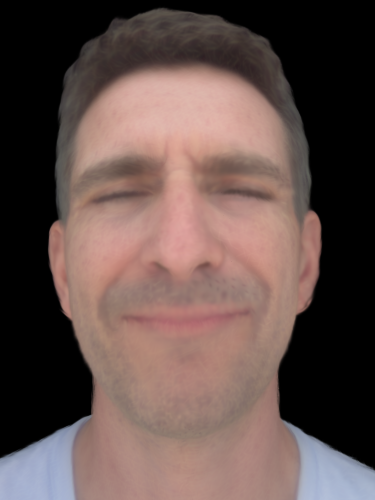}
    & \includegraphics[width=\ITWimg]{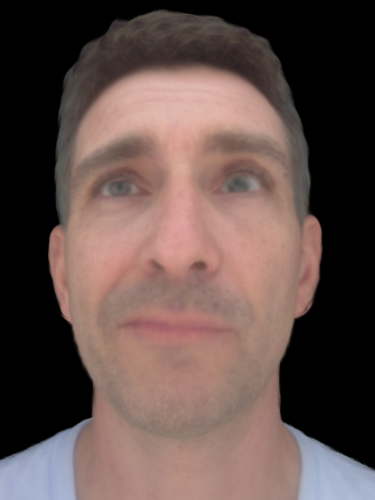}
    & \includegraphics[width=\ITWimg]{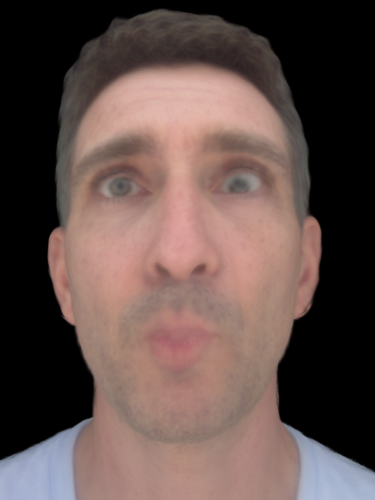}
    & \includegraphics[width=\ITWimg]{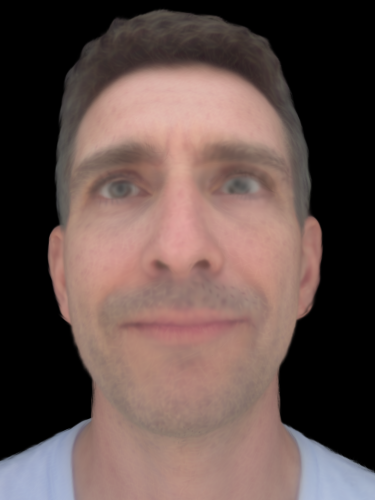} \\

    & \ITWrowlbll{\shortstack{Driving\\expression}}
    & \ITWcropimg{\includegraphics[width=\ITWimg]{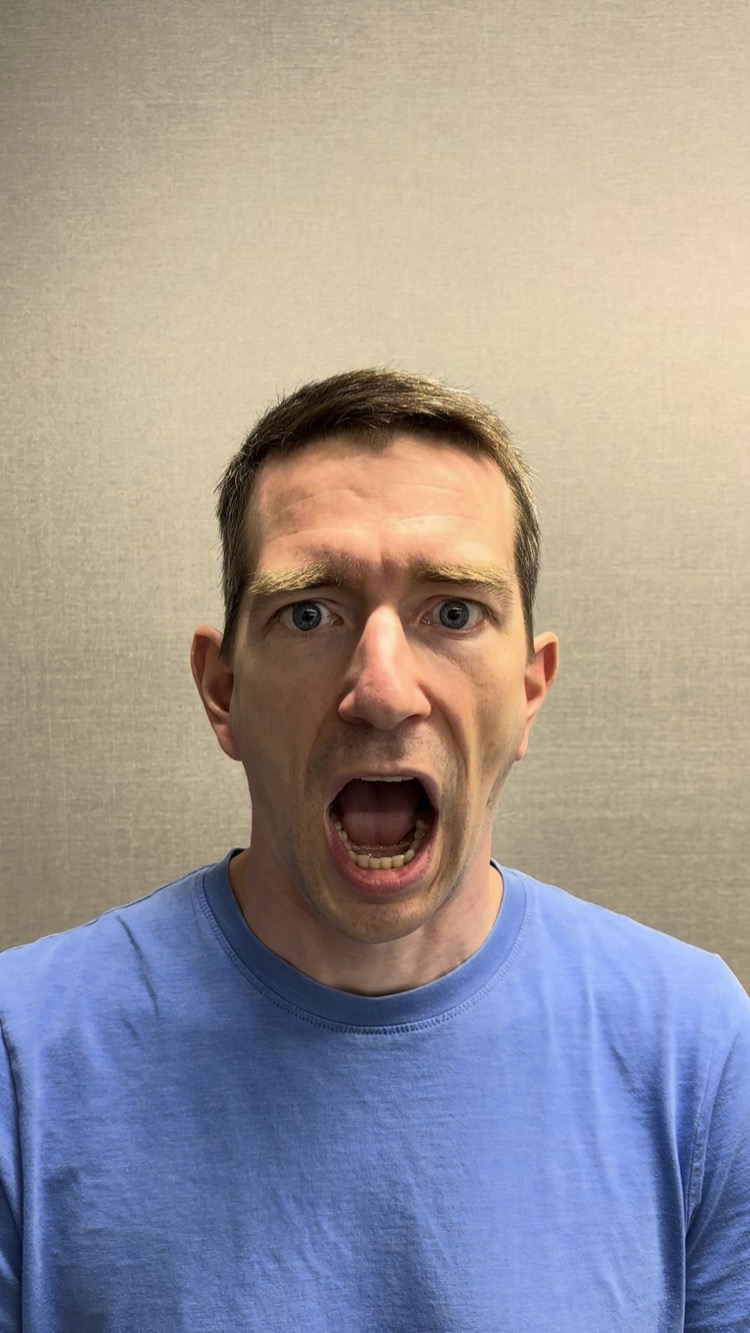}}
    & \ITWcropimg{\includegraphics[width=\ITWimg]{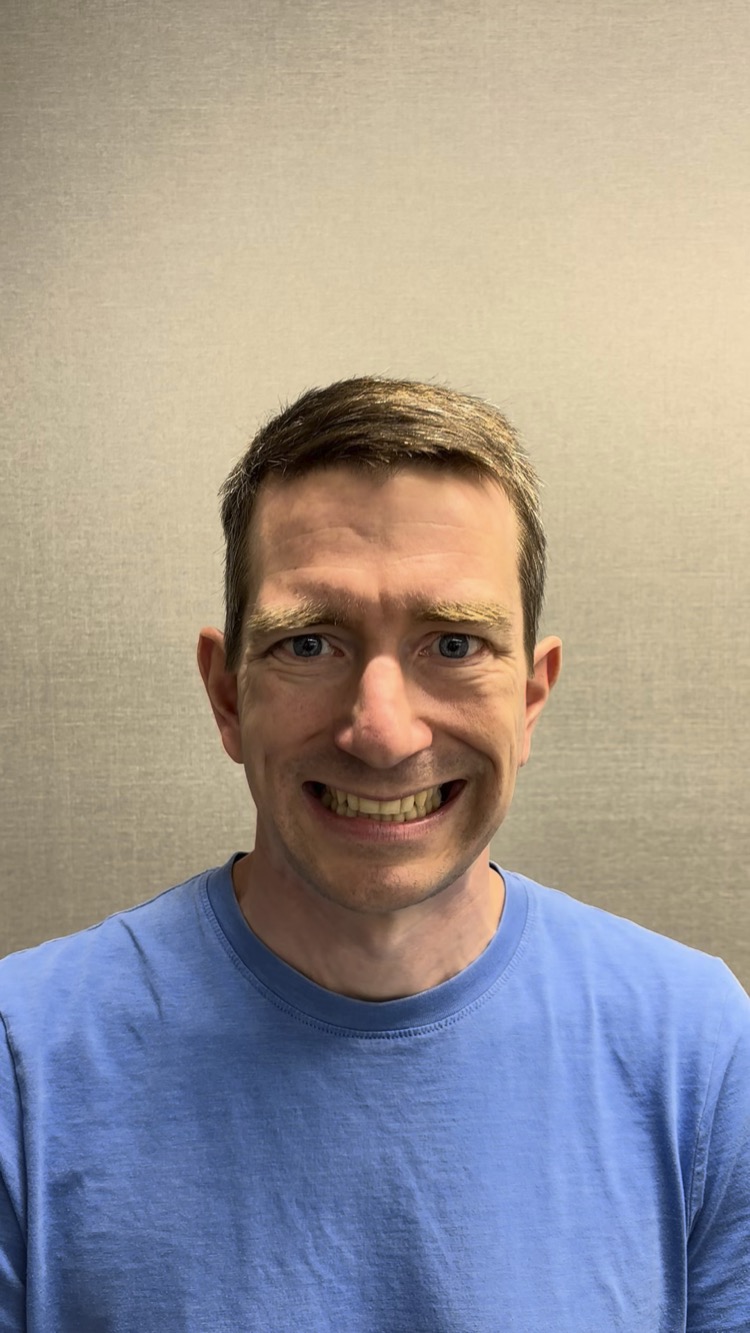}}
    & \ITWcropimg{\includegraphics[width=\ITWimg]{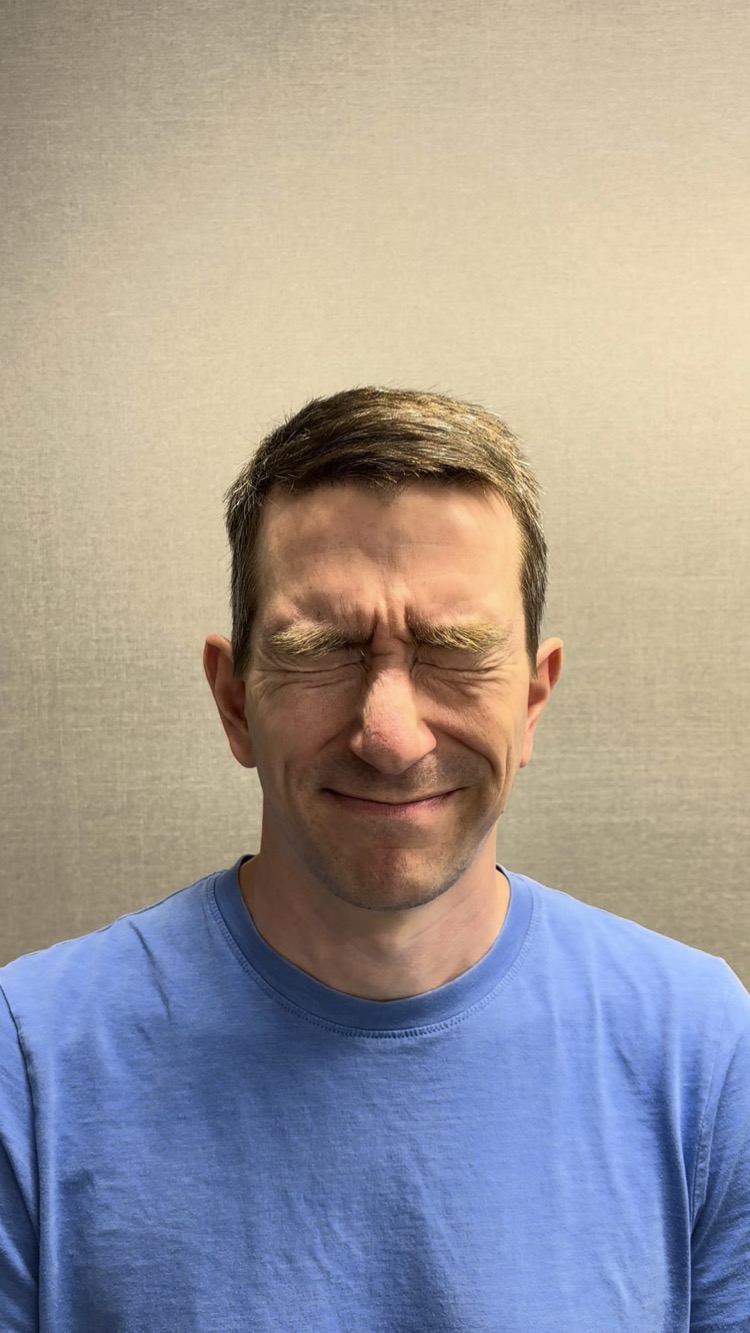}}
    & \ITWcropimg{\includegraphics[width=\ITWimg]{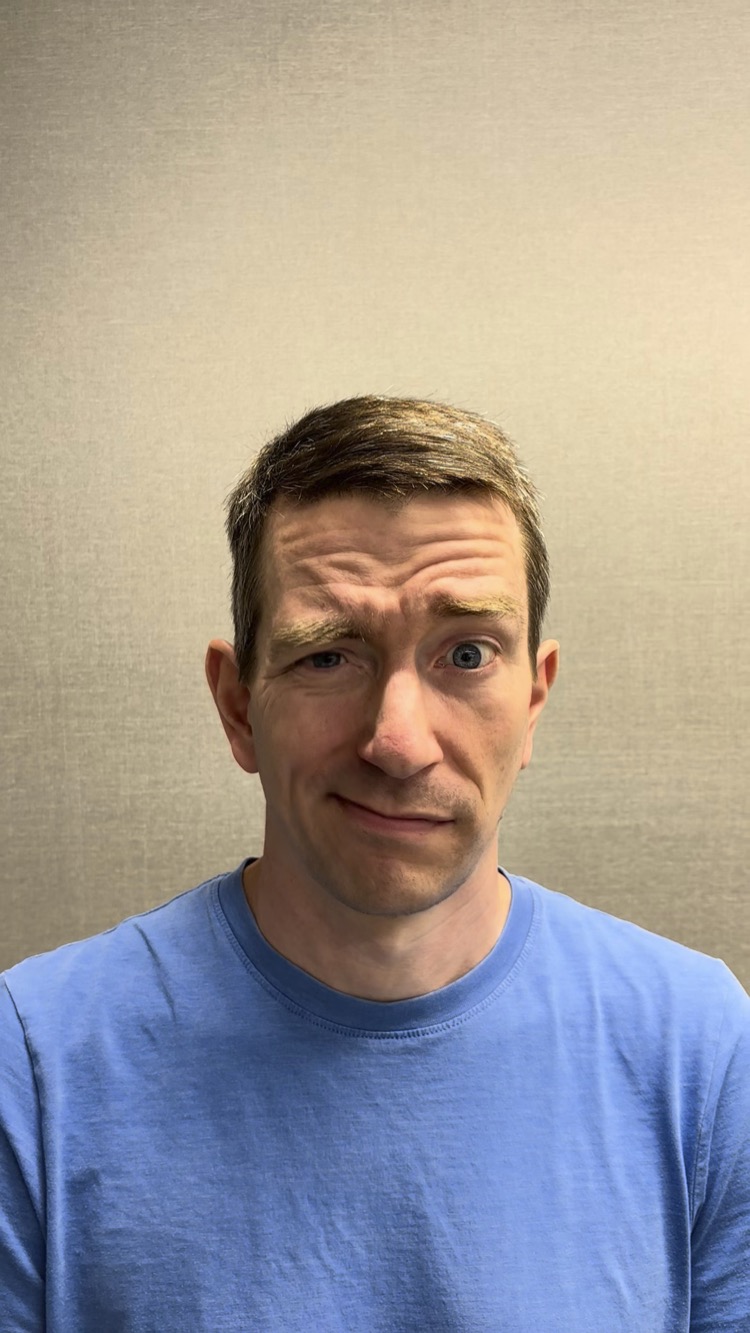}}
    & \ITWcropimg{\includegraphics[width=\ITWimg]{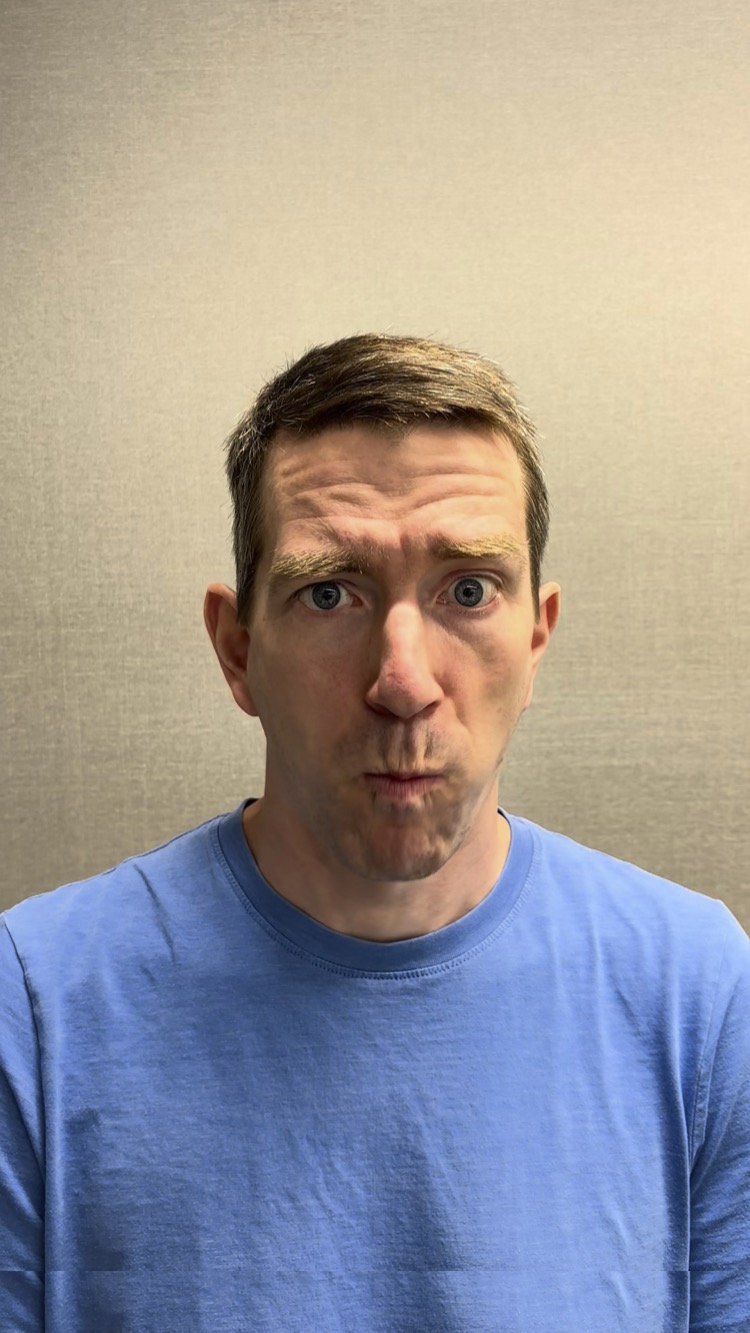}}
    & \ITWcropimg{\includegraphics[width=\ITWimg]{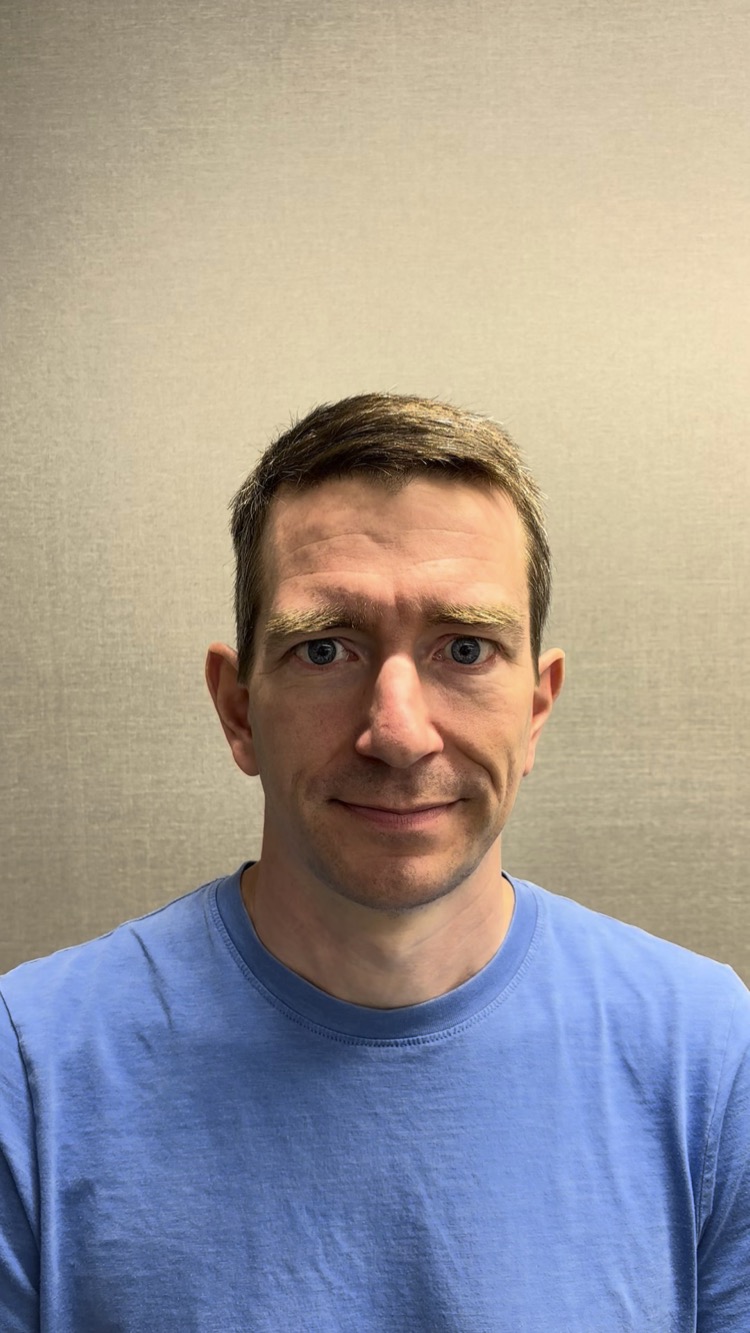}} \\
  \end{tabular}
  \caption{\textbf{In-the-wild driving image.} The bottom row shows images captured with an iPhone. We use ARKit to extract expression codes from these images and drive the avatar from the input image of the same subject (left column).}
  \label{fig:inthewild}
\end{figure*}

In this section, we evaluate our algorithm on in-the-wild data. Fig.~\ref{fig:inthewild} shows an example of self-reenactment for our algorithm compared with Avat3r and FastGHA. We extract the expression codes using ARKit from a video recorded with an iPhone 15. The video frames are shown in the bottom row. Then we captured an image of the same subject that we used as input for the identity generation (left column). We use the models trained for one input view for this experiment. All the models receive the exact same inputs. The self-reenactment rows show the animated avatar with the expressions from the bottom row and the identity from the left column. Since ARKit~\cite{arkit} is robust to in-the-wild data, and we abstract the driving expression into expression codes, the algorithms can handle in-the-wild reference expressions without issues. 
Our model reconstructs identity with a high level of detail from a single input image, whereas FastGHA \cite{ji2026fastgha} and Avat3r \cite{kirschstein2025avat3r} exhibit substantially greater degradation in fine detail and identity preservation.
Our model relies on the HeadsUp prior, which was trained on studio data with specific camera configurations and focal lengths. Jointly adapting the prior and our model to more diverse capture settings is an interesting direction for future work.


\subsection{Cross-Identity Retargeting} \label{sec:suppXID}

\begin{figure*}
\centering
\makebox[\textwidth]{%
  \makebox[\dimexpr\textwidth/6\relax][c]{\small Input}%
  \makebox[\dimexpr\textwidth/6\relax][c]{\small \shortstack{Driving\\expression}}%
  \makebox[\dimexpr\textwidth/6\relax][c]{\small Avat3r}%
  \makebox[\dimexpr\textwidth/6\relax][c]{\small FastGHA}%
  \makebox[\dimexpr\textwidth/6\relax][c]{\small Ours}%
  \makebox[\dimexpr\textwidth/6\relax][c]{\small Ours$_{+\text{BAN}}$}%
}\\[1pt]
\includegraphics[width=\textwidth, trim=0 615 1000 0, clip]{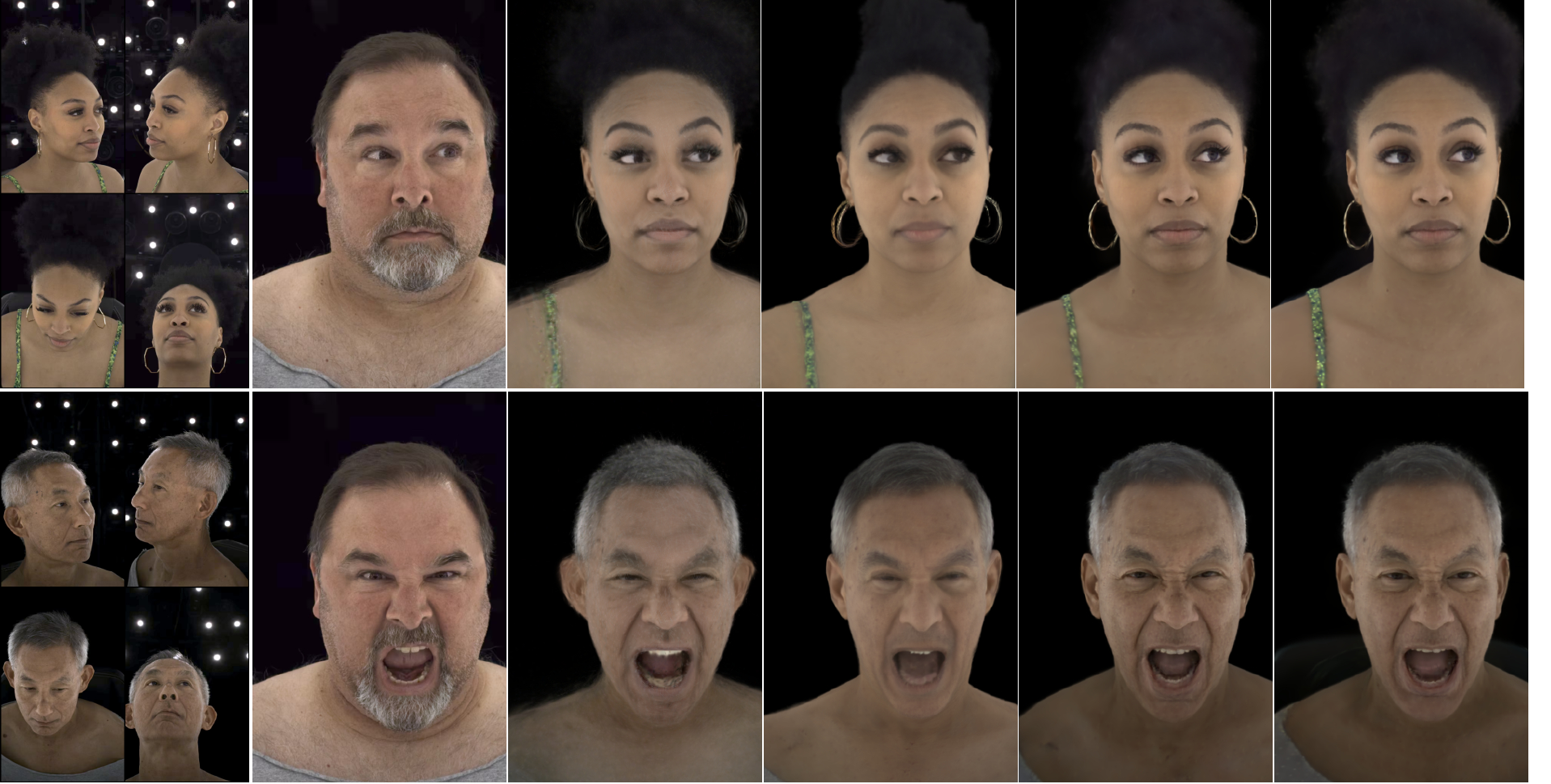}
\caption{\textbf{More results for cross-identity retargeting.} For each pair, the identity input is shown in the first column, and the expression is driven by the subject in the second column.}
\label{fig:more_results_xid}
\end{figure*}

Fig.~\ref{fig:more_results_xid} shows more results for cross-identity retargeting. Similar to Fig.~\ref{fig:ava256_cross_identity} in the main paper, our method shows better expression transfer across different identities than Avat3r and FastGHA (more extreme expression and expression-dependent details like wrinkles).

\subsection{Comparison with UIKA}
\label{app:uika}
We compare our method with the concurrent UIKA~\cite{wu2026uika}, using the authors' official code and checkpoint finetuned on Ava-256 to convergence. As shown in Tab.~\ref{tab:uika}, our method obtains the best AKD and PSNR using half the Gaussians of UIKA (65K vs.\ 134K), while UIKA is slightly better on LPIPS and SSIM. Our animation stage is also faster than UIKA on every device where both can run. UIKA cannot run on the iPhone ANE because of the large LBS matrix multiplications.

\begin{table}[h!]
\centering
\setlength{\tabcolsep}{4pt}
\begin{tabular}{lrccccc}
\toprule
Method & \#G & LPIPS$\downarrow$ & SSIM$\uparrow$ & AKD$\downarrow$ & CSIM$\uparrow$ & PSNR$\uparrow$ \\
\midrule
UIKA~\cite{wu2026uika} & 134K & \textbf{0.19} & \textbf{0.79} & 3.9 & \textbf{0.67} & 21.3 \\
Ours & \textbf{65K} & 0.20 & 0.76 & \textbf{3.1} & \textbf{0.67} & \textbf{22.2} \\
\bottomrule
\end{tabular}

\vspace{6pt}
\begin{tabular}{lrccc}
\toprule
 & & A100 & iPhone 17 Pro (ANE) & iPhone 17 Pro (GPU) \\
Method & \#G & ms / FPS & ms / FPS & ms / FPS \\
\midrule
UIKA~\cite{wu2026uika} & 134K & 2.13 / 470 & -- & 9.99 / 100 \\
Ours & \textbf{65K} & \textbf{1.92 / 522} & \textbf{1.45 / 690} & \textbf{5.10 / 196} \\
\bottomrule
\end{tabular}
\vspace{5pt}
\caption{\textbf{Comparison with UIKA on Ava-256} with 4 input views. Top: reconstruction quality. Bottom: animation runtime. \#G is the number of Gaussians. UIKA cannot run on the ANE because of the large LBS matrix multiplications.}
\label{tab:uika}
\end{table}

\end{document}